\documentclass{IEEEtaes}

\jvol{XX}
\jnum{XX}
\jmonth{XXXXX}
\paper{1234567}
\pubyear{2022}
\doiinfo{TAES.2022.Doi Number}

\usepackage[pdftex]{graphicx} 
\usepackage{amsmath} 
\usepackage{algorithmic}
\usepackage{color}
\usepackage{array} 
\usepackage[caption=false,font=footnotesize]{subfig} 
\usepackage{dblfloatfix}
\usepackage{url} 

\usepackage{microtype}
\usepackage[utf8]{inputenc} 
\usepackage[T1]{fontenc}    
\usepackage{paralist} 
\usepackage{marvosym} 
\usepackage{bm} 
\usepackage{bbm} 
\usepackage{amssymb}   
\usepackage{mathrsfs} 
\usepackage{tabularray} 
\UseTblrLibrary{booktabs}       
\usepackage{tabularx} 
\usepackage[flushleft]{threeparttable} 
\usepackage{colortbl} 
\usepackage{makecell} 
\usepackage{multirow} 
\usepackage{nicefrac} 
\usepackage[dvipsnames]{xcolor} 
\usepackage{soul} 
\setul{0.5ex}{0.3ex}

\definecolor{citecolor}{RGB}{34,139,34}
\usepackage{cleveref}  
\crefformat{figure}{Fig.~#2#1#3}
\crefformat{equation}{Eq.~(#2#1#3)}
\crefformat{table}{Table~#2#1#3}
\crefformat{section}{Section~#2#1#3}
\crefformat{algorithm}{Algorithm~#2#1#3}
\crefformat{appendix}{Appendix #2#1#3}
\crefformat{subsection}{subsection~#2#1#3}
\usepackage{cellspace}
\usepackage[numbers,sort&compress]{natbib} 

\usepackage{xspace} 

\usepackage{pifont} 
\newcommand{\xmark}{\ding{53}}%

\definecolor{Gray}{rgb}{0.9,0.9,0.9}
\definecolor{LightCyan}{rgb}{0.88,1,1}
\newcolumntype{a}{>{\columncolor{Gray}}c}
\newcolumntype{b}{>{\columncolor{white}}c}

\begin{document}

\setlength{\abovedisplayskip}{.5\baselineskip} 
\setlength{\belowdisplayskip}{.5\baselineskip} 

\title{DenoDet: Attention as Deformable Multi-Subspace Feature Denoising for Target Detection in SAR Images}


\author{YIMIAN DAI}
\affil{Nanjing University of Science and Technology, Nanjing, China}

\author{MINRUI ZOU}
\affil{Nanjing University of Posts and Telecommunications, Nanjing, China}

\author{YUXUAN LI}
\affil{Nankai University, Tianjin, China}

\author{XIANG LI}
\affil{Nankai University, Tianjin, China}

\author{KANG NI}
\affil{Nanjing University of Posts and Telecommunications, Nanjing, China}

\author{JIAN YANG}
\affil{Nanjing University of Science and Technology, Nanjing, China}

\receiveddate{Manuscript received XXXXX 00, 0000; revised XXXXX 00, 0000; accepted XXXXX 00, 0000.\\
This work was supported by the National Natural Science Foundation of China (62301261, 
62101280, 
62206134, 
62361166670), 
the Natural Science Foundation of Jiangsu Province (BK20210588), and the China Postdoctoral Science Foundation under Grant (2021M701727, 2023M731781).
}

\corresp{{\itshape The first two authors contributed equally to this work. (Corresponding author: Kang Ni and Jian Yang)}.}

\authoraddress{
  Yimian Dai and Jian Yang are with the PCA Lab, Key Lab of Intelligent Perception and Systems for High-Dimensional Information of Ministry of Education, and Jiangsu Key Lab of Image and Video Understanding for Social Security, School of Computer Science and Engineering, Nanjing University of Science and Technology.
(e-mail: \href{mailto:yimian.dai@gmail.com}{yimian.dai@gmail.com}, \href{mailto:csjyang@mail.njust.edu.cn}{csjyang@mail.njust.edu.cn}).\\
Minrui Zou and Kang Ni are with Nanjing University of Posts and Telecommunications, Nanjing, China. Kang Ni is also with Key Laboratory of Radar Imaging and Microwave Photonics, Nanjing University of Aeronautics and Astronautics, Ministry of Education, Nanjing, China. (e-mail: \href{mailto:traveler\_wood@163.com}{traveler\_wood@163.com}, \href{mailto:tznikang@163.com}{tznikang@163.com}). \\
Yuxuan Li and Xiang Li are with VCIP, CS, Nankai University. 
(e-mail:
  \href{mailto:yuxuan.li.17@ucl.ac.uk}{yuxuan.li.17@ucl.ac.uk};
  \href{mailto:xiang.li.implus@nankai.edu.cn}{xiang.li.implus@nankai.edu.cn}).
}

\supplementary{Color versions of one or more of the figures in this article are available online at \href{http://ieeexplore.ieee.org}{http://ieeexplore.ieee.org}.}

\markboth{YIMIAN DAI ET AL.}{DENODET}
\maketitle


\begin{abstract}
Synthetic Aperture Radar (SAR) target detection has long been impeded by inherent speckle noise and the prevalence of diminutive, ambiguous targets.
While deep neural networks have advanced SAR target detection, their intrinsic low-frequency bias and static post-training weights falter with coherent noise and preserving subtle details across heterogeneous terrains.
Motivated by traditional SAR image denoising, we propose DenoDet, a network aided by explicit frequency domain transform to calibrate convolutional biases and pay more attention to high-frequencies, forming a natural multi-scale subspace representation to detect targets from the perspective of multi-subspace denoising.
We design TransDeno, a dynamic frequency domain attention module that performs as a transform domain soft thresholding operation, dynamically denoising across subspaces by preserving salient target signals and attenuating noise.
To adaptively adjust the granularity of subspace processing, we also propose a deformable group fully-connected layer (DeGroFC) that dynamically varies the group conditioned on the input features.
Without bells and whistles, our plug-and-play TransDeno sets state-of-the-art scores on multiple SAR target detection datasets. 
The code is available at \url{https://github.com/GrokCV/GrokSAR}.
\end{abstract}

\vspace{-1\baselineskip}

\section{Introduction} \label{sec:introduction}

S{\scshape ynthetic} Aperture Radar (SAR), an active microwave imaging sensor, excels in all-weather, round-the-clock surveillance due to its ability to penetrate atmospheric obstructions such as clouds, rain, snow, and fog \cite{Measurement2020PolSarSegmentation}. 
This unique capability makes SAR indispensable for a wide range of applications, spanning from military reconnaissance \cite{IJRS2020ApproachOil} and precision guidance to civil domains such as maritime search and rescue, resource exploration, and disaster management \cite{JSTARS2022TechniqueOilSar}.
Consequently, the precise detection of targets from complex terrestrial environments using SAR imagery is of great practical importance \cite{JISRS2019SarHybird}.

\begin{figure*}[htbp]
  \centering
    \vspace{-1\baselineskip}
  \subfloat[MSAR dataset]{
    \includegraphics[width=.225\textwidth]{
      "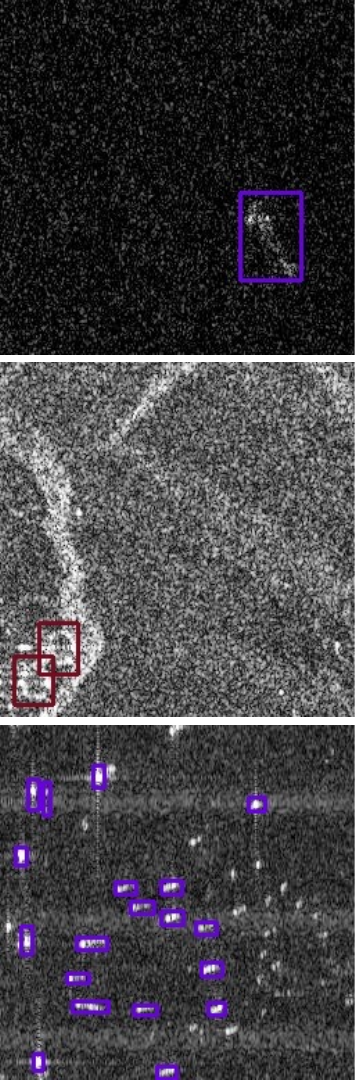"}
      \label{subfig:Gallery-MSAR}
  }
  \subfloat[SAR-AIRCraft-1.0 dataset]{
    \includegraphics[width=.225\textwidth]{
      "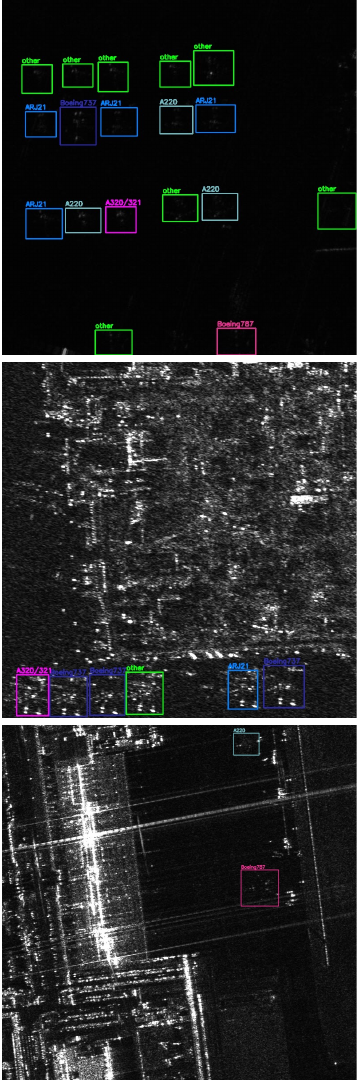"}
      \label{subfig:Gallery-AIRCraft}
  }
  \subfloat[AIR-SARShip-1.0 dataset]{
    \includegraphics[width=.225\textwidth]{
      "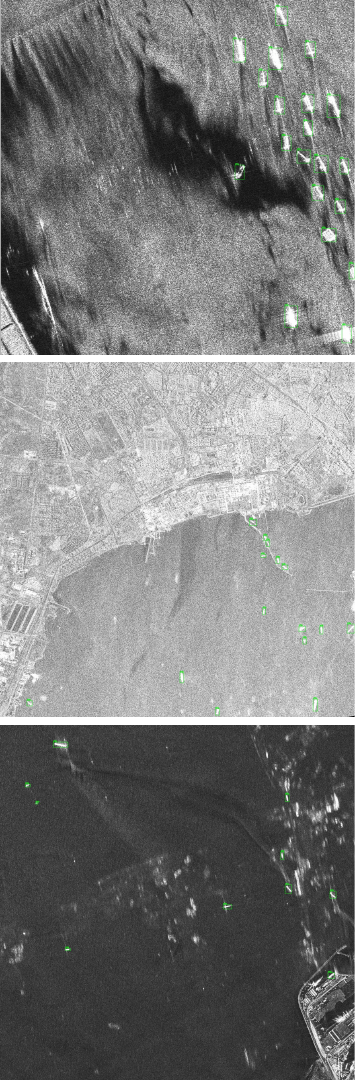"}
      \label{subfig:Gallery-AIRCraft}
  }
  \subfloat[SARDet-100K dataset]{
    \includegraphics[width=.225\textwidth]{
      "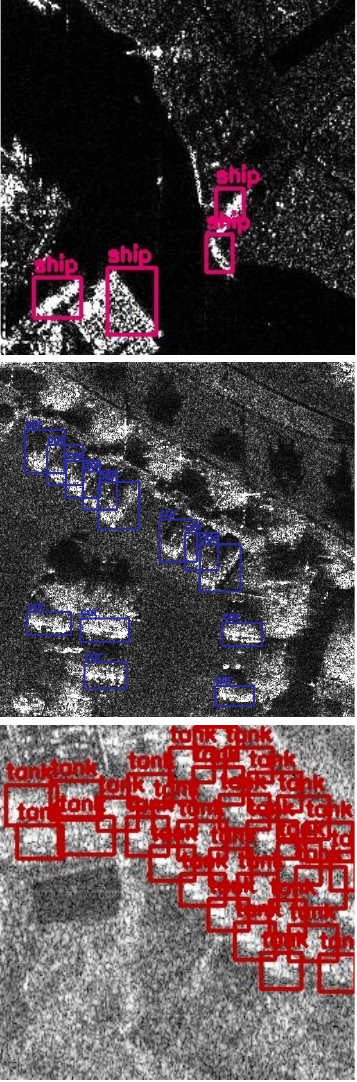"}
      \label{subfig:Gallery-AIRCraft}
  }
  \caption{Representative imagery far SAR target detection: (a) MSAR dataset \cite{chen2022large}, (b) SAR-AIRcraft-1.0 dataset \cite{zhirui2023sar}, (c) AIR-SARShip-1.0 dataset \cite{xian2019air}, and (d) SARDet-100K dataset \cite{arXiv2024SARDet}. These images illustrate two inherent challenges: 1) \textbf{the presence of unavoidable speckle noise}, and 2) \textbf{most of targets being relatively small entities}, possessing ambiguous intrinsic features that render them susceptible to noise interference.}
  \label{fig:gallery}
\end{figure*}

Although SAR provides numerous advantages, as depicted in Fig.~\ref{fig:gallery}, it also introduces distinct challenges for target detection, especially when compared to generic object detection tasks in optical images. Specifically:
\begin{itemize}
  \item[1)] \textbf{Speckle Noise Interference:} As SAR is a coherent imaging system, its images intrinsically contain unavoidable speckle noise, overlaid on the targets, significantly increasing the difficulty of target detection and identification \cite{GRSM21PixelStatistical}.
  \item[2)] \textbf{Challenges with Small Targets:} Limited by the spatial resolution of imaging, the targets of interest mostly appear as small targets in SAR images, with indistinct intrinsic features that are more easily interfered by noise \cite{TGRS22AnchorFreeSAR}.
\end{itemize}


\vspace*{-1\baselineskip}
\subsection{Proir Works on Target Detection in SAR Images}


Traditional SAR image target detection methods primarily leverage structural, grayscale, and texture features of targets.
Geometric attributes such as area, aspect ratio, rectangularity, principal axis orientation, moment of inertia, and fractal dimension play a vital role in boosting the accuracy and efficiency of ship detection.
Pioneering studies have optimized the extraction of these attributes, refining methods for closed-contour representation \cite{EJRS2022GreyWolfLSM} and enhancing orientation angle estimates through bifurcated approaches. Notably, Akbarizadeh \cite{TGRS2012NewWavelet} developed a robust texture recognition algorithm for SAR images by leveraging introducing novel statistical features like Kurtosis Wavelet Energy (KWE) and Skewness Wavelet Energy (SWE) to improve segmentation under speckle noise.

Recently, deep learning methods have revolutionized the field of SAR target detection \cite{IETIP2019ChangSarDetection}, offering superior performance over traditional approaches.
One of the primary challenges in SAR target detection is the presence of speckle noise, which can significantly degrade the detection accuracy.
To mitigate this issue, researchers have proposed various techniques, such as using multi-scale feature representations \cite{TGRS22Multikernel}, incorporating context information \cite{TRGS2024SRT-Net}, and employing soft thresholding \cite{TGRS2023SoftThresholding}. 
For instance, Zhang et al. \cite{TGRS2023SoftThresholding} introduced an oriented ship detection network based on soft thresholding and context information, which demonstrated improved performance in complex scenes. Similarly, Jia \textit{et al.} \cite{TGRS2024FastDetection} proposed a fast progressive ship detection method that combines the advantages of traditional non-DL methods and DL approaches. Other studies have focused on developing more robust features, such as the kurtosis wavelet energy feature \cite{TGRS2012NewWavelet} and the lacunarity cue \cite{TGRS2022SaliencySar}. These advancements have significantly enhanced the robustness of SAR target detection methods to SAR noise, leading to improved detection rates.

Detecting small targets in SAR images is another significant challenge, as they often have limited spatial extent and can be easily confused with clutter.
To address this issue, researchers have proposed various techniques, such as using shallow feature enhancement networks with dilated convolution, employing multi-kernel-size feature fusion \cite{TGRS22Multikernel}, and incorporating scattering region topology \cite{TRGS2024SRT-Net}.
For instance, Pan \textit{et al.} \cite{TRGS2024SRT-Net} introduced a scattering region topology network (SRT-Net) for oriented ship detection, which demonstrated improved performance in detecting small targets.
Other studies have focused on developing more effective loss functions, such as the dual Euclidean distances loss \cite{TGRS2023SidelobAware}, and incorporating scene characteristic learning \cite{TGRS2023SemisupervisedSar}. 
Additionally, techniques such as semisupervised learning \cite{TGRS2023SemisupervisedSar} and pruning \cite{TGRS2024SARGap} have been explored to improve the efficiency and effectiveness of SAR target detection methods. 
Recent studies have also investigated the use of deep learning-based methods for oil spill detection \cite{Geo2022osdes_net,JSTARS2022TechniqueOilSar} and corona detection \cite{TPD2021CoronaDetection}. These advancements have improved the detection rates of small targets in SAR images.

Despite significant advances in SAR target detection, a fundamental tension at the core of feature extraction hampers the attainment of high accuracy. In our opinion, current methods, while effective to an extent, act as mere stopgaps rather than addressing the fundamental issues underlying SAR image processing. The limitations of existing approaches manifest in several critical areas:
\begin{enumerate}
  \item \textbf{Insufficient Noise Suppression:} CNNs inherently perform multi-stage spatial filtering, capturing target features while simultaneously attempting spatial denoising. However, the coherent speckle noise in SAR images, intermixed with targets features, is difficult to segregate in the original spectral domain, often leading to the inadvertent removal of essential details.
  \item \textbf{Spectral Bias in Deep Learning:} Theoretical analysis \cite{arXiv22SpectralBias} reveals that deep networks exhibit a spectral bias towards low-frequency components, which predominantly represent background or larger objects. This bias results in the blurring of edges and the loss of fine details, particularly detrimental for the detection of small or subtle targets.
  \item \textbf{Static Network Weights:} A critical limitation arises from the static nature of network weights post-training, which do not adjust adaptively to varying terrains or noise conditions in SAR imagery. This rigidity often leads to either over-smoothing or under-filtering, thereby obfuscating pivotal target details that are essential for accurate detection.
\end{enumerate}


Given the challenges outlined above, a naive question might arise: Can we simply denoise the SAR images before performing target detection?
However, extensive research \cite{NIPS2022RestorationforDet} has demonstrated the ineffectiveness of such approaches, as the complete separation of coherent speckle noise from target features in the image domain is an inherently ill-posed problem. The removal of noise inevitably leads to the loss of crucial target details, compromising the accuracy of subsequent detection tasks.


\vspace*{-1\baselineskip}
\subsection{Motivation}

Faced with the dilemma of balancing noise suppression and subtle feature preservation, we are compelled to ask: Rather than attempting to suppress speckle noise at the image level, \textit{can we shift the focus of denoising to the feature level within the target detection framework}?
In other words, can we \textit{\textbf{design a plug-and-play feature denoising module that can be seamlessly integrated into a target detection framework}}, thereby improving the detection accuracy of SAR targets? 


Our answer is affirmative. 
To address the challenge of accurately detecting targets in SAR images, we propose \textbf{DenoDet}, a network that \textbf{improves the performance of SAR target detection from the perspective of multi-subspace feature denoising in the frequency domain}.
Our motivation stems from the desire to enable the network, through explicit frequency domain transforms, to prioritize the preservation of high-frequency information associated with small target features while calibrating the convolutional network's inherent bias towards low-frequency components. 
The key insight is that in the frequency domain, different frequencies correspond to structures of varying scales, naturally forming a multi-scale subspace representation.
This allows us to explicitly adopt adaptive operations for subspaces of different frequencies.
Furthermore, compared to spatial filtering, frequency domain transforms, such as the discrete cosine transform (DCT), are global in nature, enabling features in the frequency domain to capture a broader context beyond local receptive fields.


Driven by this motivation, we propose a plug-and-play frequency domain attention module, \textbf{TransDeno}, which can be interpreted as a dynamic feature denoising process across multiple subspaces in the frequency domain.
The core idea behind TransDeno is that \textit{\textbf{attentional feature refinement can be viewed as a dynamic soft thresholding denoising process}}, selectively preserving effective target signals while filtering out redundant noise components.
Specifically, given a feature, TransDeno employs a 2D DCT to map the feature representation into the frequency domain and decomposes it into multiple subspaces, each representing information within a specific frequency range.
For each frequency component, our approach applies a dynamic element-wise soft thresholding based on its full context within the corresponding subspace.

A critical aspect of our approach lies in the determination of the optimal number of frequency-domain subspaces.
Considering the significant variations in scene content across different images, \textit{we argue that the number of subspaces should be adaptively determined for each individual image}.
To address this issue, we propose the \textbf{Deformable Group Fully Connected (DeGroFC) layer}, which dynamically ascertains the optimal number of subspaces for each image by intelligently modulating the grouping within fully connected layers. 
This ensures a tailored response to the unique spectral characteristics of each SAR image, effectively capturing and exploiting the inherent spectral diversity across different SAR scenes for enhanced feature representation and improved target detection performance.




In summary, our contributions can be categorized into \textbf{FOUR} main aspects:
\begin{itemize}
  \item Our DenoDet offers \textbf{a novel interpretation of attention mechanisms} as dynamic soft threshold shrinkage operations, akin to transform domain denoising.
  \item We propose TransDeno, a grouped attention module that performs multi-subspace element-wise soft thresholding in the transform domain, enhancing both computational efficiency and detection performance.
  \item We propose the DeGroFC layer, which dynamically selects the number of subspaces in TransDeno, tailored to enhance target signal retention and suppress noise in a content-aware manner.
  \item Our DenoDet achieves state-of-the-art performance on multiple SAR target detection datasets, demonstrating its effectiveness. Notably, it attains the rank of No. 1 on the leaderboard of SARDet-100K, the largest SAR target detection dataset.
\end{itemize}

\section{Related Work} \label{sec:related}

\begin{figure*}[htbp]
  \centering
  \includegraphics[width=.99\textwidth]{"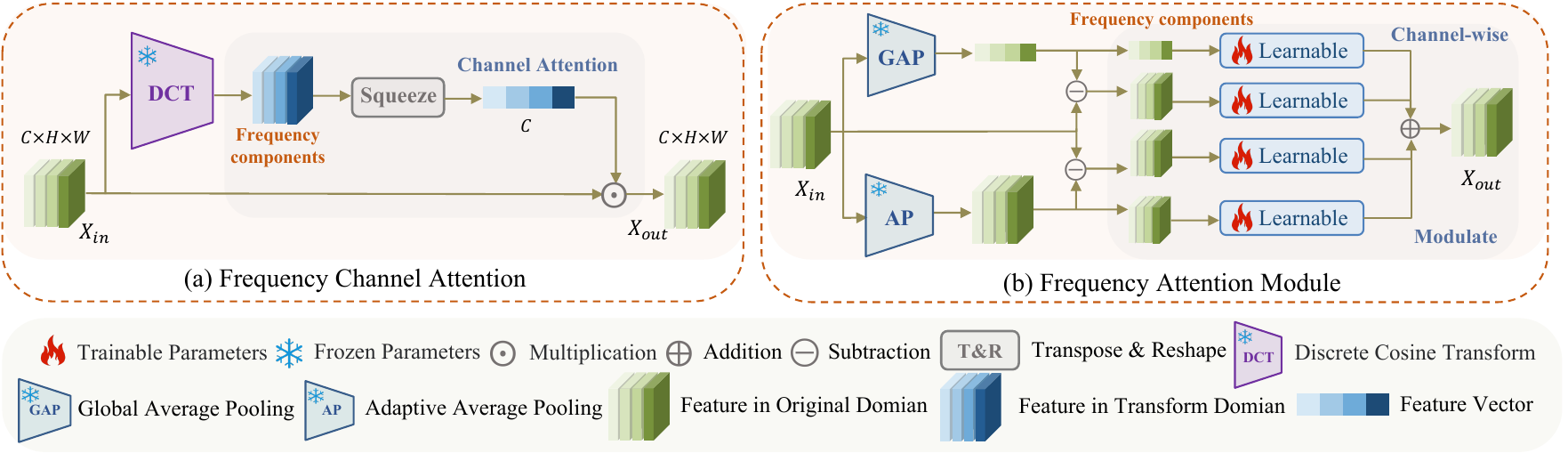"}
  \caption{Existing frequency attention modules: (a) Frequency Channel Attention~\cite{ICCV21FCA} and (b) Frequency Attention Module~\cite{AAAI23DualDomainAttention}.
  }
  \label{fig:TransDeno}
\end{figure*}

\subsection{Target Detection in SAR Images}
SAR target detection is a critical and yet complex task due to its intricate imaging mechanisms and the pervasive speckle noise. Traditional approaches typically involve despeckling as a preliminary step, but this can inadvertently remove valuable information, compromising target detection efficacy. Despite improvements in visual quality, the suitability of despeckling as a precursor to detection warrants further investigation.

Deep learning methods for SAR target detection often borrow from advancements in optical imaging, with a focus on refining network architectures to bolster detection capabilities. For example, Yang et al. \cite{yang2023super} integrated a lightweight feature extraction network with a slim bidirectional feature pyramid into the YOLOv5 framework to efficiently capture SAR target characteristics. PVT-SAR \cite{zhou2022pvt} harnessed a Pyramid Vision Transformer (PVT) \cite{wang2021pyramid} to obtain multiscale global features through self-attention. Domain adaptation algorithms \cite{zhao2022feature, shi2021unsupervised} have been tailored to address discrepancies between training and testing SAR datasets during deep feature learning.

Recent initiatives have leveraged frequency-domain insights to enhance SAR target detection, inspired by successes in general vision tasks. Li et al. \cite{TGRS2022Multidimensional} introduced a multi-domain deep learning network that utilizes the spatial and complementary features from a polar Fourier transform for ship detection in SAR imagery. Zhang et al. \cite{TGRS2023FAM} presented a frequency attention module to mitigate sea clutter by incorporating frequency-domain data.


Our DenoDet model aligns with the objective of capitalizing on frequency-domain learning, but diverges from existing methods in several unique ways:
\begin{enumerate}
    \item \textbf{Reimagined Attention Mechanisms:} Unlike existing frequency domain attention modules, DenoDet reimagines attention mechanisms with a focus on transform domain denoising.
    \item \textbf{Deformable Grouping in FC Layer:} Another innovation in our DenoDet model is the introduction of the DeGroFC layer. This layer dynamically adjusts the grouping based on the input features, \textit{a concept not previously explored in literature}. 
\end{enumerate}

\subsection{Attention Mechanism}

The attention mechanism effectively enhances neural representations for diverse tasks, allowing models to dynamically prioritize distinct segments of input data, facilitating the processing of intricate visual information. The Squeeze-and-Excitation (SE) block, which learns inter-channel correlations and recalibrates channel-wise features, has been introduced to refine these representations further \cite{hu2018squeeze}. The Selective Kernel Network (SKNet) incorporates a dynamic selection mechanism, enabling neurons within a Convolutional Neural Network (CNN) to adjust their receptive fields adaptively \cite{li2019selective}.

To augment channel attention, incorporating long-range spatial interactions has proven beneficial. For instance, the Gather-Excite Network (GENet) employs depth-wise convolution to aggregate and redistribute feature responses across a broad spatial range. The Convolutional Block Attention Module (CBAM) sequentially integrates channel and spatial attention to enhance feature processing \cite{woo2018cbam}. The Large Selective Kernel Network (LSKNet) optimizes its expansive spatial receptive field for diverse object scenarios in remote sensing \cite{li2023large}. Additionally, Deformable Convolutional Networks (DCN) have pioneered the use of deformable convolutional layers, enabling flexible adaptation of the convolutional kernel's receptive field to the spatial context of targets \cite{CVPR19DCNv2}.
Our proposed TransDeno module extends the concept of spatial attention, distinguishing itself by operating in the frequency domain, employing a dynamic soft-thresholding denoising process, and featuring a deformable group fully connected layer---unlike the primarily 2D convolution-focused methods previously mentioned.

Recent studies have integrated high-frequency information into attention mechanisms. For example, Cui \emph{et al.} created a frequency attention module that selectively emphasizes informative frequency components \cite{AAAI23DualDomainAttention}. Qin \emph{et al.}'s Frequency Channel Attention Networks (FcaNet) adapt channel attention mechanisms to the frequency domain \cite{ICCV21FCA}. 
Our TransDeno module aligns with the objective of enhancing feature representation through advanced blocks but differs in several ways:
\begin{enumerate}
    \item \textbf{Frequency Domain Recalibration:} Unlike FcaNet that focuses primarily on channel-specific recalibration, TransDeno organizes features by frequency, adjusting them in the frequency domain, and converting them back to the spatial domain using IDCT.
    \item \textbf{Spatial Attention with Grouped FC Layer:} While FcaNet applies 2D DCT for channel-specific attention, our TransDeno module emphasizes spatial attention. We employ a group fully connected layer to limit cross-frequency spatial interactions.
    \item \textbf{Dynamic Modulation with DeGroFC Layer:} DeGroFC layer allows the module to adapt to the varying complexity of features, while existing modules only operate with fixed groups.
\end{enumerate}



\section{Methods} \label{sec:method}

\begin{figure*}[htbp]
  \centering
  \includegraphics[width=.99\textwidth]{"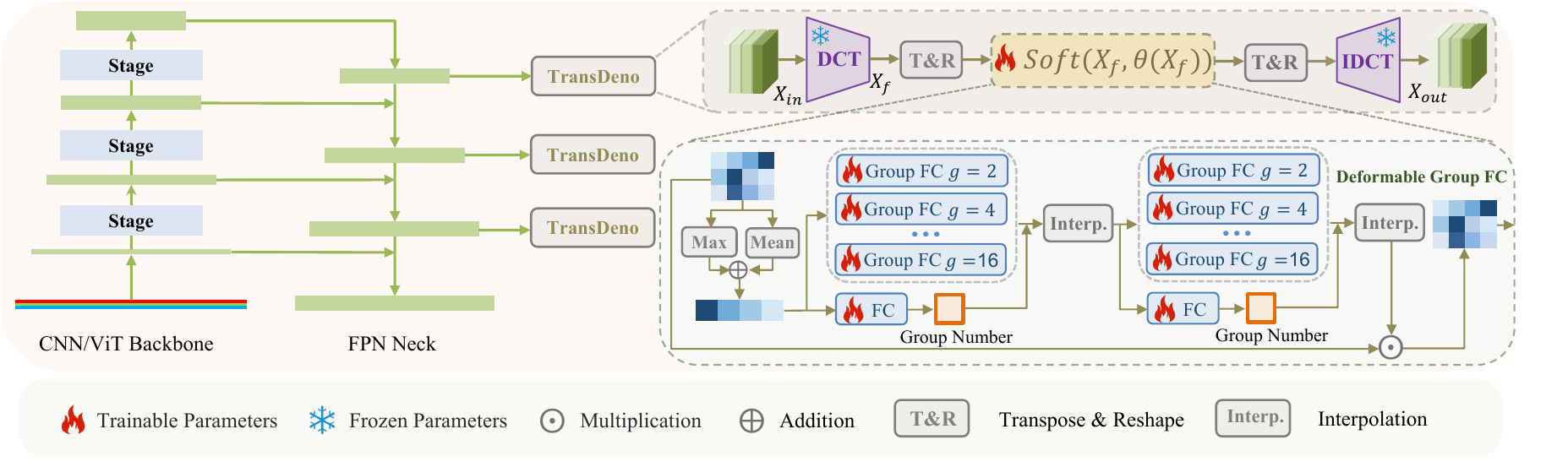"}
  \caption{Conceptual illustration of the proposed TransDeno module.
  At its core, the TransDeno module is envisioned as a dynamic soft thresholding denoising operation within the transform domain. Central to this architecture is the proposed deformable group fully connected layer, depicted in the light blue region, which dynamically partitions the feature space into subsets for independent processing.
  This design enables the module to adapt to the diverse frequency distribution patterns of different targets, enhancing the module's ability to capture fine-grained target details.
  }
  \label{fig:denodet-arch}
\end{figure*}

\subsection{Revisiting Attention as Dynamic Soft Thresholding}

Soft thresholding is a fundamental technique in signal and image processing for noise reduction \cite{TIP00AdaptiveThresholding}. It diminishes low-intensity noise by employing a thresholding function that progressively reduces signal values toward zero while retaining the signal's significant elements. The mathematical representation of soft thresholding is:
\begin{equation}
\operatorname{soft}(x, \theta) = \operatorname{sign}(x) * \operatorname{max}(0, |x| - \theta),
\label{eq:soft-thresholding}
\end{equation}
where $x$ denotes the input signal value. $\operatorname{sign}(x)$ indicates the sign of $x$ and $|x|$ is the absolute value, and $\theta$ represents the threshold.

For more effective noise suppression, signals are often converted to a specific transform domain, like the wavelet or Fourier domain, to more clearly differentiate between signal features and noise. Following soft thresholding, the signal is reverted to its original domain. This process is encapsulated by:
\begin{equation}
\hat{\mathbf{x}}=\widetilde{\mathcal{W}}\left(\operatorname{soft}\left(\mathcal{W}\left(\mathbf{x}\right), \theta\right)\right),
\end{equation}
where $\mathcal{W}$ and $\widetilde{\mathcal{W}}$ represent the forward and inverse transformations, respectively, with $\widetilde{\mathcal{W}} \circ \mathcal{W} = \mathcal{I}$ denoting the identity operator.

The advent of attention mechanisms has been transformative in machine learning, enabling models to selectively concentrate on specific segments of input data. These mechanisms allocate varying weights to data components, emphasizing those most pertinent to prediction tasks. For an intermediate feature map $\mathbf{X} \in \mathbb{R}^{C \times H \times W}$, attention mechanisms induce the following transformation \cite{ICPR21ATAC}:
\begin{equation}
  \mathbf{X}^{\prime}=\mathbf{G}\left(\mathbf{X}\right) \odot \mathbf{X},
  \label{Eq:OverallAttention}
\end{equation}
where $\odot$ represents element-wise multiplication and $\mathbf{G}(\mathbf{X}) \in \mathbb{R}^{C \times H \times W}$ is a weight map generated by the attention gating module $\mathbf{G}$. The gating module's output is contingent on the entire feature map $\mathbf{X}$. Consequently, at a specific position $(c, i, j)$, we can reformulate Eq. (\ref{Eq:OverallAttention}) in a scalar form as:
\begin{align}
  \mathbf{X}^{\prime}_{[c, i, j]} = \mathbf{G}(\mathbf{X})_{[c, i, j]} \cdot \mathbf{X}_{[c, i, j]}
                                  = g_{c, i, j}\left(\mathbf{X}\right) \cdot \mathbf{X}_{[c, i, j]}, 
  \label{Eq:ElemAttention}
\end{align}
where $g$ symbolizes a complex gating function. Given $(c, i, j)$, the function $g$ aggregates relevant feature context, generates attention weights, and selectively applies these weights to $\mathbf{X}_{[c, i, j]}$. Actually, Eq. (\ref{Eq:ElemAttention}) can be reformulated as
\begin{equation}
  \mathbf{X}^{\prime}_{[c, i, j]} = \operatorname{sign}(\mathbf{X}_{[c, i, j]}) * \operatorname{max}(0, |\mathbf{X}_{[c, i, j]}| - \theta_{[c, i, j]}),
  \label{eq:attention-soft}
\end{equation}
where $\theta_{[c, i, j]} = (1 - g_{c, i, j}\left(\mathbf{X}\right)) \cdot |\mathbf{X}_{[c, i, j]}|$ is the dynamic threshold conditioned on the input feature.


\textbf{Discussion:} A comparison of Eq. (\ref{eq:soft-thresholding}) and Eq. (\ref{eq:attention-soft}) reveals that \textbf{both attention mechanisms and soft thresholding denoising function as shrinkage operators}. They only differ in that the threshold $\theta$ in soft thresholding is generally constant, whereas in attention mechanisms, it is a dynamic value dependent on the input. 
The key benefit of modeling attention mechanisms as soft thresholding denoising is that \textbf{it allows recasting the feature modulation problem as a feature denoising task}. This enables borrowing and adapting techniques from the extensive field of image denoising to inform the design of novel attention modules. In particular, since soft thresholding methods for SAR image denoising are predominantly based on transform domains, attention mechanisms can be extended to operate in transformed spaces as well. This promises further performance gains for attention modules, as elaborated in the next subsection.

\subsection{DenoDet Architecture}


The architecture of our proposed DenoDet is depicted in Fig.~\ref{fig:denodet-arch}. It builds upon a generic object detection network, enhanced by the integration of our TransDeno module. This module executes dynamic soft threshold denoising within the transform domain of the feature maps. TransDeno is designed as a versatile, plug-and-play module that can theoretically be incorporated at various stages of the network, including the Backbone, Neck, or Head. In our framework, we position it between the feature refinement neck and the detection head, capitalizing on the refined feature maps that contain rich contextual information, which is advantageous for detection tasks. This placement also permits the detection head to utilize more distinctive representations.



\begin{figure}[htbp]
  \centering
    \vspace{-1\baselineskip}
  \subfloat[2D DCT filters]{
    \includegraphics[width=.235\textwidth]{
      "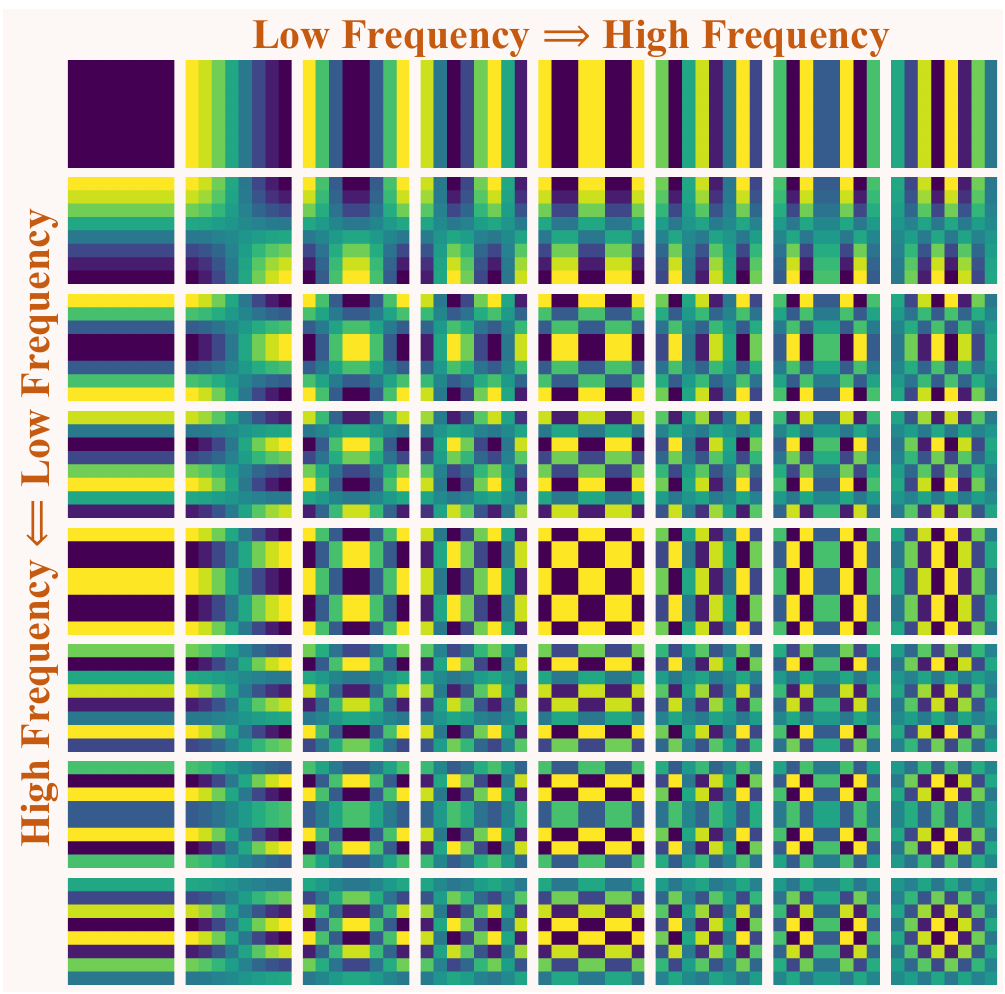"}
      \label{subfig:dct-filters}
  }
  \subfloat[Frequency Subspaces]{
    \includegraphics[width=.235\textwidth]{
      "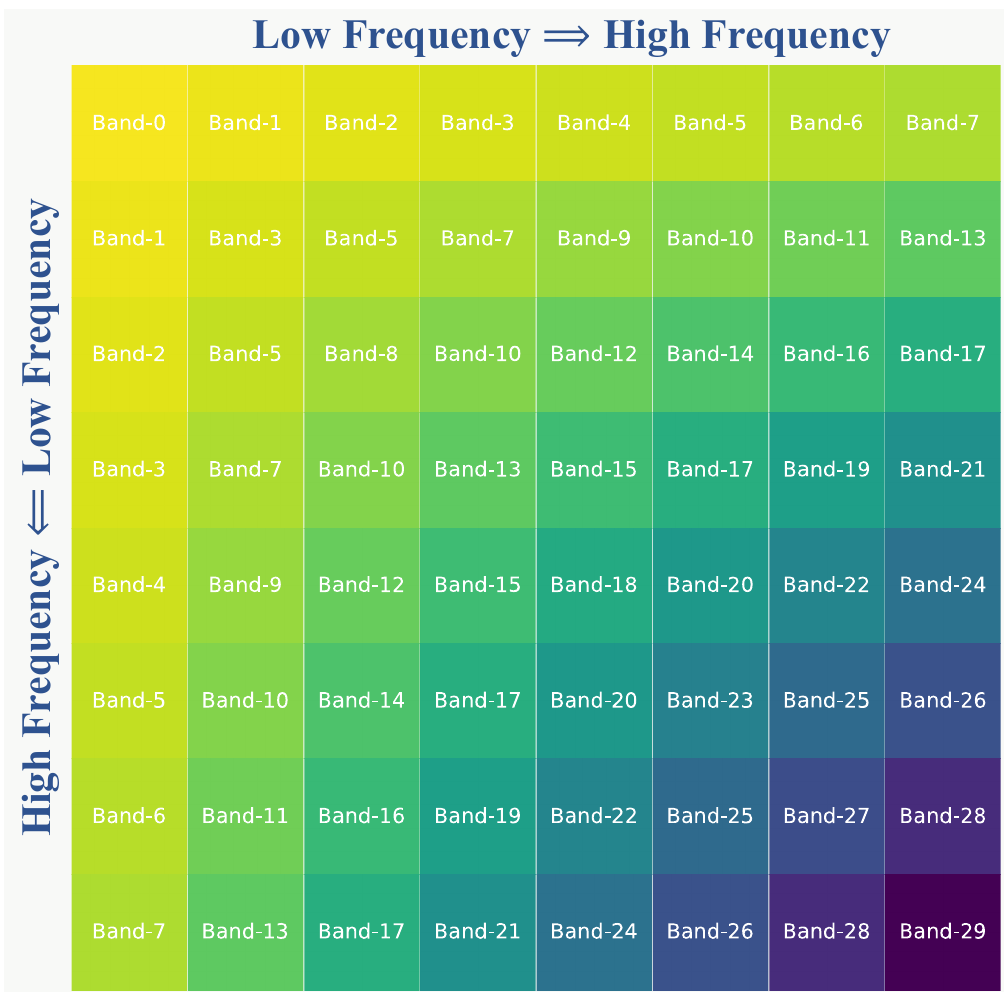"}
      \label{subfig:frequency-bands}
  }
  \caption{Spectral decomposition of image features using 2D DCT filters.
  (a) Illustration of 2D DCT filters employed in our module. (b) Schematic representation of the frequency domain feature partitioning. Semantically meaningful target features and irrelevant noise often occupy different frequency subspaces in the transform domain.}
  \label{fig:dct}
\end{figure}

As illustrated in Fig.~\ref{fig:denodet-arch}, the TransDeno module comprises a 2D Discrete Cosine Transform (DCT) for forward transformation, dynamic soft threshold denoising, followed by an inverse 2D DCT. For a feature map $\mathbf{M}\in{\mathbb{R}^{C \times H \times W}}$, the 2D DCT forward transformation $\mathcal{W}$ is mathematically expressed as:
\begin{equation}
\begin{array}{l}
\mathbf{m}_{c,i,j}
 = \mathop \sum \limits_{h = 0}^{H - 1} \mathop \sum \limits_{w = 0}^{W - 1} {\textbf{M}_{c,h,w}}{\rm{cos}}\left( {\frac{{\pi i}}{H}\left( {h + \frac{1}{2}} \right)} \right){\rm{cos}}(\frac{{\pi j}}{W}\left( {w + \frac{1}{2}} \right)).\;\;\\
s.t.\quad h, i \in \left\{ {0,1,\cdots,H - 1} \right\}, w, j \in \left\{ {0,1,\cdots,W - 1} \right\}.
\end{array}
\end{equation}
Here, $\mathbf{m}$ represents the transformed feature map in the frequency domain, with $h$ and $w$ indicating the spatial coordinates in the original feature map, and $i$, $j$ indexing the frequency components in the DCT spectrum.
The depiction of 2D DCT filters is presented in Fig.~\ref{fig:dct} (a), illustrating a spectral hierarchy where features are extracted in an ascending order of frequency. 
As explicated in Fig.~\ref{fig:dct} (b), these spectral features can be methodically partitioned into distinct subspaces, each corresponding to a specific frequency range. 

In the spectral domain, the transformation of spatial information into frequency components via DCT reveals a distinctive pattern: low-frequency components, indicative of global spatial characteristics, originate in the upper-left quadrant of the spectrum, progressing toward high-frequency details in the lower-right. 
Our research endeavors to harness this hierarchical spectrum by decomposing the frequency domain features and allocating them into dedicated subspaces that correspond to discrete spectral bands, thereby sharpening our focus on relevant signals.


To facilitate this targeted analysis, we unfolds two-dimensional (2D) feature maps into one-dimensional (1D) vectors. This transformation ensures that the full breadth of spectral information, spanning the gamut from low to high frequencies, is preserved and made accessible within each individual subspace. The mathematical formalism for this transformation is given by:
\begin{equation}
\widetilde{\mathbf{m}} = \mathcal{R}(\mathbf{m}): {\mathbb{R}^{C \times H \times W}} \rightarrow {\mathbb{R}^{HW \times C}},
\end{equation}
where $\mathcal{R}$ represents the reshaping operation that converts the feature map $\mathbf{m}$ into a rearranged matrix $\widetilde{\mathbf{m}}$.
Conversely, the inverse operation is defined as $\mathbf{m} = \widetilde{\mathcal{R}}(\widetilde{\mathbf{m}}): {\mathbb{R}^{HW \times C}} \rightarrow {\mathbb{R}^{C \times H \times W}}$.

Our DenoDet integrates an attention mechanism that assimilates global semantic information to modulate signals in the spectral domain. For an input feature map $\mathbf{m}\in{\mathbb{R}^{C \times H \times W}}$ in the frequency domain, we derive a frequency spectrum map by executing average and maximum pooling operations along the channel dimension:
\begin{equation}
\begin{array}{l}
\widetilde{\mathbf{S}}_{\operatorname{f}} = \mathcal{R} \left( \mathcal{P}_{\operatorname{avg}}(\mathbf{m}) + \mathcal{P}_{\max }(\mathbf{m}) \right),
\end{array}\
\end{equation}
where $\mathcal{P}_{\operatorname{avg}}(\mathbf{m}), \mathcal{P}_{\max }(\mathbf{m}) \in \mathbb{R}^{1 \times H \times W}$ yield the frequency spectrum feature statistics.

To dynamically compute element-wise thresholds $\widetilde{\mathbf{\Theta}}$, we fed frequency-domain feature maps into an attention block, which enable networks to optimize the signal strength of low-frequency and high-frequency information by the input data.
\begin{equation}
\widetilde{\mathbf{\Theta}}\left(\widetilde{\mathbf{m}}\right) = \left(1 - \sigma \left( \mathcal{D}_{g^{\prime}} \left( \delta \left( \mathcal{D}_g ( \widetilde{\mathbf{S}}_{\operatorname{f}} ) \right) \right) \right) \right) \odot \widetilde{\mathbf{m}},
\label{eq:Theta}
\end{equation}
where $\mathcal{D}_{g}$ epitomizes the proposed deformable group fully connected layer with with $g$ groups, the particulars of which will be elaborated in the subsequent section.
$\delta$ denotes the ReLU function, and $\sigma$ is sigmoid function. 
This mechanism empowers the network to fine-tune the signal strength of both low-frequency and high-frequency information based on the input data. 
Finally, the process of our TransDeno module is succinctly encapsulated as:
\begin{equation}
\begin{array}{l}
\mathbf{M'}
 = \widetilde{\mathcal{W}}\left(
 \widetilde{\mathcal{R}} \left( 
\operatorname{soft}(\widetilde{\mathbf{m}}, \widetilde{\mathbf{\Theta}}\left(\widetilde{\mathbf{m}}\right)) 
 \right)
 \right),
\end{array}\
\end{equation}
where $\mathbf{M'}$ is output feature map of TransDeno module. $\widetilde{\mathcal{W}}$ denotes DCT inverse transformation defined as:
\begin{equation}
\begin{array}{l}
\mathbf{M}_{c,h,w}
 = \mathop \sum \limits_{i = 0}^{H - 1} \mathop \sum \limits_{j = 0}^{W - 1} {\textbf{m}_{c,i,j}}{\rm{cos}}\left( {\frac{{\pi i}}{H}\left( {h + \frac{1}{2}} \right)} \right){\rm{cos}}(\frac{{\pi j}}{W}\left( {w + \frac{1}{2}} \right)),\;\;\\
s.t.\quad h,i \in \left\{ {0,1,\cdot\cdot\cdot,H - 1} \right\},\;w,j \in \left\{ {0,1,\cdot\cdot\cdot,W - 1} \right\}.
\end{array}\
\end{equation}

\begin{figure}[htbp]
  \centering
  \includegraphics[width=.475\textwidth]{"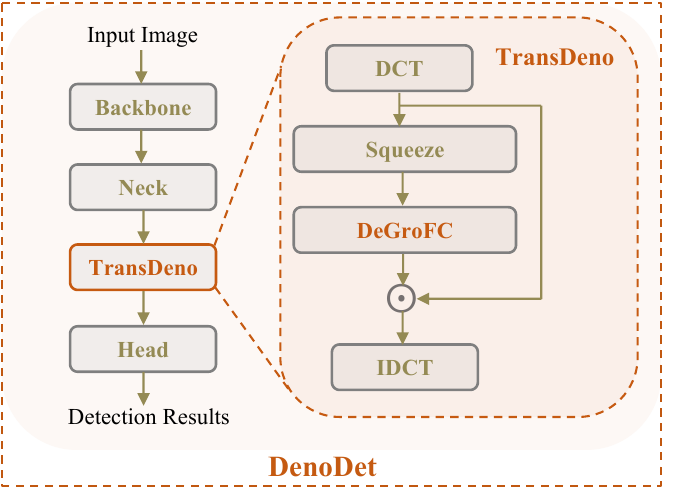"}
  \caption{A simplified flowchart of our DenoDet framework.
  }
  \label{fig:flowchart}
\end{figure}

To facilitate a better understanding of our proposed method, we provide a simplified flow diagram in Fig.~\ref{fig:flowchart}, which outlines the interplay among our contributions: DenoDet, TransDeno, and DeGroFC.
In a nutshell, DeGroFC is a part of TransDeno, which in turn is an essential module within the DenoDet framework.

\textbf{Discussion:} 
A key distinction of this work compared to previous methods is the application of attention mechanisms in the frequency domain. In contrast to spatial domain features which have high flexibility in arrangement, frequency domain features follow a strict low-to-high frequency layout based on the bandwidth, as illustrated in Fig. \ref{fig:dct} (b). This makes frequency representations more amenable to modeling by fixed-position, fixed-weight network structures like fully connected layers. Specifically, the network weights depend only on frequency rather than spatial location. This design enables easier learning of frequency distribution patterns, allowing better capture of fine-grained target details.
Additionally, semantically meaningful target features and irrelevant noise often occupy different frequency subspaces in the transform domain. Hence, the introduction of frequency domain features also facilitates separating targets from noise, which will be elaborated in the next subsection.

\subsection{Deformable Group FC Layer}

The quintessence of our DenoDet is the segmentation of the spectral domain into a series of distinct subspaces for further processing. A pertinent challenge that arises in this context is the determination of the optimal number of subspaces---a decision that is critical for the performance of our model. A rudimentary strategy for establishing the number of groups is to rely on the detection performance on a validation set, as evidenced by the experiments detailed in Table~\ref{tab:deformable}. However, this way does not account for the variability in the granularity required for subspaces across different images, suggesting the necessity of a more adaptive approach.

To address this issue, we propose the Deformable Group Fully Connected (DeGroFC) layer, as illustrated in Fig. \ref{fig:denodet-arch}, a pivotal addition to our architecture that confers upon it the capability to dynamically ascertain the optimal number of subspaces for processing each image.
This selection is not arbitrary; it is meticulously tailored to the unique spectral signatures inherent within each SAR image.
The DeGroFC layer is a conceptual leap, inspired by the flexibility of Deformable Convolutional Networks \cite{CVPR19DCNv2}, and incorporates a channel attention mechanism that fluidly modulates the grouping of fully connected layers.

Imagine a scenario where we have $k$ possible group configurations pre-defined in our design space. To navigate this space, a dedicated fully connected layer is employed to generate $k$ coefficients, serving as the navigational compass that guides the selection of the grouping strategy. This process is mathematically expressed as:
\begin{equation}
{K}_{i}= {\cal F}^{H*W \to k}(\widetilde{\mathbf{S}}_{\operatorname{f}}),
\end{equation}
where $\cal{F}$$^{H*W \to k}$ symbolizes the transformation performed by the fully connected layer.
As can be seen, the dynamic number of groups $K_i$ in DeGroFC primarily depends on the input squeezed feature $\widetilde{\mathbf{S}}_{\text{f}}$, in which different semantic contents will lead to different values of $K_i$, allowing adaptation to varying feature distributions across images.

An ideal ${K}_{i} \in [0,k-1]$ represents the $i$-th value among the set of $k$ coefficients. In this paper, $k$ is set to 4, and the coefficients represent the group number, \textit{i.e.}, $[2, 4, 8, 16]$.
In fact, ${K}_{i}$ is typically fractional. 
Upon deriving the coefficients ${K}_{i}$, we engage in a rounding operation to determine the offsets, which ultimately dictate the chosen strategy:
\begin{equation}
\begin{array}{l}
{O_i^p} = {K_i} - \lfloor {{K_i}} \rfloor,\\
{O_i^q} =  \lceil {{K_i}} \rceil  - {K_i},
\end{array}\
\end{equation}
where ${O_i^p}$ and ${O_i^q}$ represent two complementary sets of offset values. These offsets are instrumental in performing bilinear interpolation, and we process them through a softmax function to embed a degree of non-linearity essential for the model's adaptability.

Given an input feature $\mathbf{x}$, for the $k$ different grouping strategies in the group fully connected layers, our DeGroFC layer can be formulated as follows:
\begin{equation}
\mathcal{D}\left( \mathbf{x} \right) = \sum\limits_{i = 1}^k \mathcal{G}_{\lfloor{K_i}\rfloor} \left( \mathbf{x} \right) * {O_i^p} + \mathcal{G}_{\lceil{K_i}\rceil} \left( \mathbf{x} \right) * {O_i^q},
\label{eq:deformable}
\end{equation}
where $\mathcal{G}_{n}$ denotes the group fully connected layer configured to operate with $n$ subspaces.
To further enhance detection performance, we apply Eq. (\ref{eq:deformable}) recursively in a two-stage pipeline as illustrated in Fig. \ref{fig:denodet-arch}. Specifically, the first application of the deformable group fully-connected layer generates an initial set of coarsely refined attention maps. We then feed these maps into a second round of attention refinement. This cascaded process allows for more fine-grained optimization of the attention maps in a progressive manner.
In essence, the DeGroFC layer is a special MLP structure, which can dynamically adapt the number of groups.

\textbf{Discussion:}
In addition to the dynamic soft thresholding process, the TransDeno module contains another attention mechanism -- the dynamic selection of group numbers. Similar to deformable convolution, our Deformable Group Fully Connected (DeGroFC) layer also utilizes bilinear interpolation to aggregate features from adjacent group fully connected layers.
The dynamism of these dual attention modules is key to enabling TransDeno to better adapt to varying feature distributions across different images, thereby improving practical application performance.





\section{Experiments} \label{sec:experiment}

\begin{figure*}[htbp]
  \centering
  \includegraphics[width=.98\textwidth]{"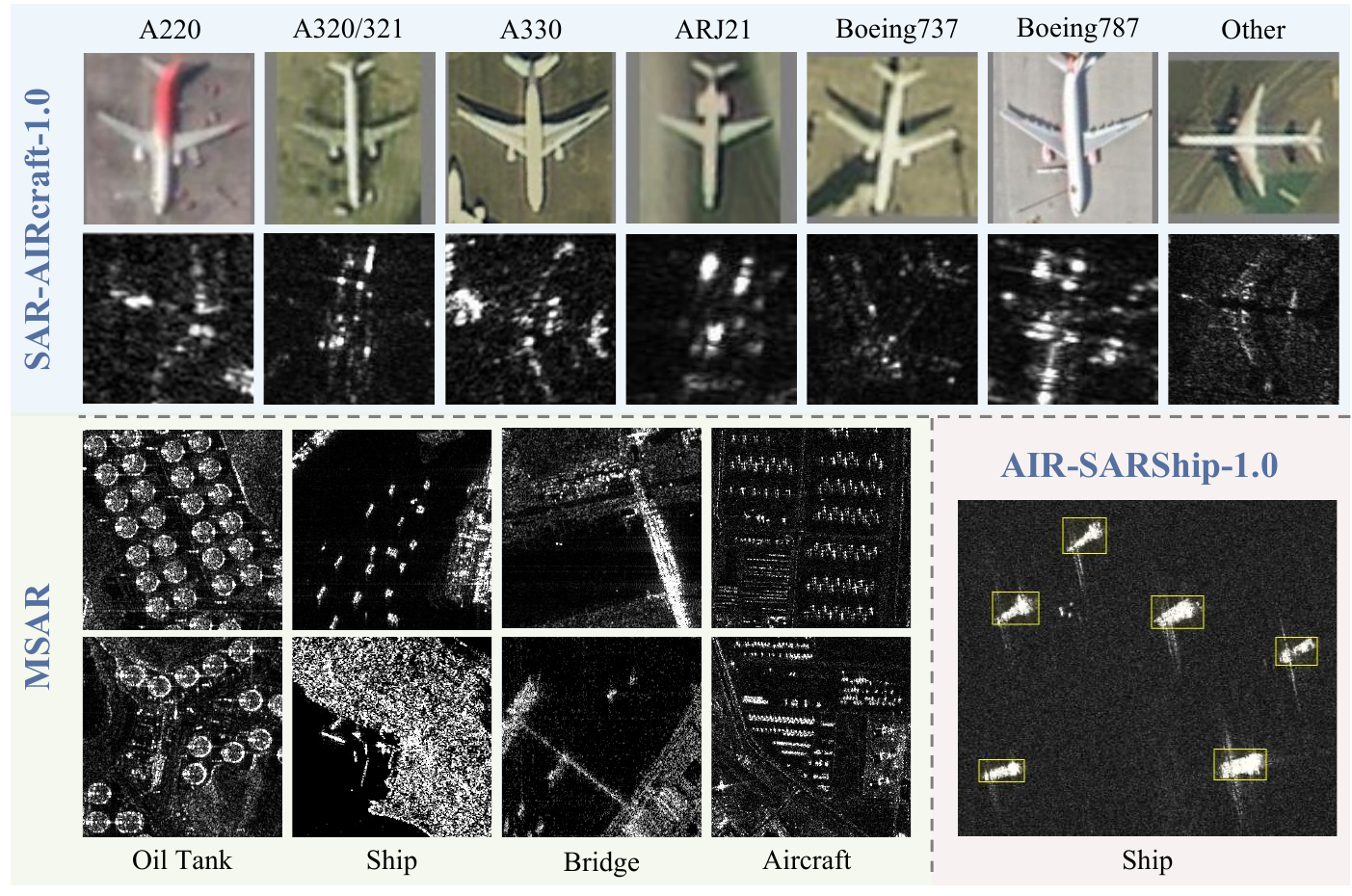"}
  \caption{Example samples from various categories in the experimental datasets: (a) SAR-AIRcraft-1.0, (b) MSAR, and (c) AIR-SARShip-1.0.
  }
  \label{fig:dataset}
\end{figure*}

\subsection{Experimental Setup} \label{subsec:setup}

\subsubsection{Datasets} \label{subsubsec:dataset}

To validate the generilization ability of the proposed method, our experimentation employed four benchmark datasets: MSAR \cite{chen2022large}, SAR-Aircraft-1.0 \cite{zhirui2023sar}, AIR-SARship-1.0 \cite{xian2019air}, and SARDet-100K \cite{arXiv2024SARDet}, as shown in Fig. \ref{fig:dataset}. Detailed information about the satellites and sensors used for these datasets is provided in Table~\ref{tab:sar_data_stat}.
\begin{enumerate}
    \item \textbf{MSAR} \cite{chen2022large}: This dataset is collected from China's first commercial SAR satellite HISEA-1. It consists of 60,396 instances from four categories: ship (39,858), oil tank (12,319), aircraft (6,368), and bridge (1,851).
    \item \textbf{SAR-Aircraft-1.0} \cite{zhirui2023sar}: A recently released dataset for fine-grained aircraft detection in SAR images. It covers seven aircraft types (A220, A320/321, A330, ARJ21, Boeing737, Boeing787, and others) with 16,463 instances in total.
    \item \textbf{AIR-SARShip-1.0} \cite{xian2019air}: This dataset is built upon 31 large-scene Gaofen-3 SAR images with VV polarization and 1m resolution. Each image has a size of about 3000$\times$3000 pixels. It contains 512 ship instances under various scenarios and weather conditions.
    \item \textbf{SARDet-100K} \cite{arXiv2024SARDet}: As the first large-scale dataset for SAR object detection, SARDet-100K includes 116,598 images and 245,653 instances from 6 categories (Aircraft, Ship, Car, Bridge, Tank, and Harbor), comparable to the COCO dataset in computer vision.
\end{enumerate}

\begin{table*}[htbp]
\renewcommand\arraystretch{1.3} 
\setlength{\tabcolsep}{1mm}
\centering
\caption{Satellites and Sensor Details for Datasets Used in Our Experiments}
\label{tab:sar_data_stat} 
\footnotesize
\begin{tabular}{>{\centering\arraybackslash}p{3.5cm}|>{\centering\arraybackslash}p{1.5cm}|>{\centering\arraybackslash}p{2cm}|>{\centering\arraybackslash}p{2.5cm}|>{\centering\arraybackslash}p{5.5cm}}

Datasets     & Res.~(m) & Band    & Polarization   & Satellites \& Sensor                      \\

\Xhline{1pt}
MSAR \cite{chen2022large}  & $\leq$ 1m & C   & HH, HV, VH, VV & HISEA-1 \\ 
SAR-AIRcraft-1.0 \cite{zhirui2023sar} & 1m   & C       & Uni-polar       & GF-3    \\
AIR\_SARShip-1.0 \cite{xian2019air}  & 1,3m          & C       & VV             & GF-3 \\
SARDet-100K \cite{arXiv2024SARDet} & 0.1$\sim$25m   & C, Ka, Ku, X & HH, HV, VH, VV, Uni-polar & Airborne SAR synthetic slic, GF-3, HISEA-1, RadarSat-2, S-1, TerraSAR-X, TanDEMX  \\
\end{tabular}
\end{table*}

\subsubsection{Implementation Details} \label{subsubsec:implementation}

To ensure consistency and fairness in our evaluation, all methods were subjected to a standardized dataset preprocessing procedure.
Models were engineered within the MMRotate framework, and a single RTX3090 GPU was employed for both the training and evaluation phases.

Images from the MSAR dataset were resized to $256 \times 256$ pixels and randomly flipped horizontally with a 50\% probability. The model underwent training for 36 epochs using the DAdaptAdam optimizer and a batch size of 64. We set the initial learning rate to 1.0 and the weight decay to 0.05, calculating the FLOPs for an image input of $256 \times 256$ pixels.

For the SAR-Aircraft-1.0 dataset, images were partitioned into $512 \times 512$ pixel sub-images with a 200-pixel overlap. Training was conducted over 12 epochs with a batch size of 32, an initial learning rate of 1.0, and a weight decay of 0.05, with FLOPs determined for a $512 \times 512$ pixel input.

In the case of the AIR-SARShip-1.0 dataset, images were also divided into $512 \times 512$ pixel sub-images with a 200-pixel overlap. The model's training spanned 72 epochs, under the same conditions as the SAR-Aircraft-1.0 dataset, and FLOPs were assessed based on the $512 \times 512$ pixel input size.

\subsection{Detection Results} \label{subsec:results}

\begin{table*}[htbp]
  \renewcommand\arraystretch{1.2}
  \footnotesize
  \centering
    \vspace{-1\baselineskip}
  \caption{Comparison with SOTA methods on the \textbf{MSAR} dataset.}
  \label{tab:msar}
  \vspace{-2pt}
  \setlength{\tabcolsep}{2.pt}
  \begin{tabular}{l|c|c|c|acccc|acccc}

  \multirow{2}{*}{Method} & \multirow{2}{*}{Pre.}    & \multirow{2}{*}{\#P $\downarrow$} & \multirow{2}{*}{FLOPs $\downarrow$} & \multicolumn{5}{c|}{Average Precision (AP'07)}  & \multicolumn{5}{c}{Average Precision (AP'12)} \\
  & & &  & \textbf{mAP} & Ship  & Airplane    & Bridge    & Oil Tank  & \textbf{mAP} & Ship  & Airplane    & Bridge    & Oil Tank \\
  \Xhline{1pt}
  \multicolumn{12}{l}{\textit{One-stage}}  \\ \hline
  FCOS~\cite{tian2019fcos} & IN     & 32.12M & 12.89G & 67.52 & 89.17 & 43.72 & 73.92  & 63.28 & 69.85 & 89.86 & 44.63 & 76.64& \underline{68.27} \\
  GFL~\cite{li2020generalized} & IN      & 32.27M & 13.08G & 67.24 & 89.21 & 42.16 & 74.05  & 63.54 & 68.91 & 90.26 & 41.14 & 76.97 & 67.28 \\
  RepPoints~\cite{yang2019reppoints} & IN      & 36.82M & 12.12G & 49.25 & 77.98 & 9.09 & 51.13  & 58.81 & 48.22 & 82.06 & 0.42 & 51.24 & 59.14 \\
  ATSS~\cite{zhang2020bridging} & IN      & 32.12M & 12.89G & 68.00 & 88.85 & 44.06 & 75.73  & 63.37 & 69.76 & 89.54 & 44.10 & 77.27 & 68.12\\
  CenterNet~\cite{zhou2019objects} & IN      & 32.12M & 12.88G & 67.18 & 89.87 & 41.05 & 74.91  & 62.92 & 69.25 & 91.10 & 41.24 & 77.05 & 67.61 \\
  PAA~\cite{kim2020probabilistic} & IN      & 32.12M & 12.89G & 63.37 & \textbf{90.29} & 22.63 & 77.01  & \underline{63.55} & 64.43 & \underline{92.24} & 20.36 & 78.29 & 66.82 \\
  PVT-T~\cite{wang2021pyramid} & IN      & 21.39M & 10.10G & 29.72 & 58.88 & 10.14 & 11.53  & 38.34 & 26.25 & 59.64 & 1.52 & 6.77 & 37.05 \\
  RetinaNet\cite{lin2017focal} & IN      & 36.39M & 13.14G & 33.71 & 63.63 & 0.88 & 23.47  & 46.84 & 33.32 & 64.97 & 0.64 & 21.08 & 46.57 \\
  TOOD~\cite{feng2021tood} & IN      & 32.03M & 12.62G & 66.58 & 89.04 & 42.78 & 71.00  & 63.49 & 68.32 & 89.88 & 42.24 & 73.91 & 67.24 \\
  DDOD~\cite{ACM2021DDOD} & IN      & 32.27M & 29.02G & 60.80 & 86.59 & 63.16 & 71.42  & 22.04 & 62.31 & 88.75 & 66.69 & 74.79 & 19.03 \\
  VFNet~\cite{zhang2021varifocalnet} & IN      & 32.72M & 12.09G & 65.31 & 87.94 & 38.80 & 71.03  & 63.49 & 66.56 & 89.21 & 36.74 & 73.12 & 67.16 \\
  YOLOF~\cite{chen2021you} & IN      & 42.41M & 6.58G & 46.39 & 69.39 & 9.09 & 58.81  & 48.25 & 44.59 & 71.72 & 0.33 & 60.07  & 46.25 \\
  YOLOX~\cite{Arixv2021YOLOX} & IN      & 8.94M & 2.13G & 68.40 & 89.89 & 63.49 & 61.46  & 58.74 & 70.42 & 90.61 & 67.06 & 64.53 & 59.49 \\
  AutoAssign~\cite{zhu2020autoassign} & IN      & 36.25M & 12.95G & \underline{69.41} & 89.71 & \underline{45.41} & \textbf{79.08}  & 63.44 & \underline{70.68} & 89.85 & \underline{45.49} & \underline{79.29}  & 68.09 \\
  \hline
  \multicolumn{14}{l}{\textit{Two-stage}}  \\
  \hline
  Faster RCNN~\cite{ren2015faster}    & IN   & 41.36M & 26.13G & 55.69 & 80.50 & 9.09 & 69.70  & 63.47 & 55.94 & 84.67 & 0.60 & 72.13 & 66.36 \\
  Grid RCNN~\cite{lu2019grid}    & IN   & 64.47M & 0.14T & 55.82 & 80.85 & 9.09 & 69.89  & 63.47 & 57.30 & 86.65 & 3.74 & 71.99 & 66.83 \\
  ConvNeXt~\cite{CVPR2022ConvNeXt} & IN      & 45.06M & 26.39G & 56.52 & 81.34 & 63.57 & 72.07  & 9.09 & 57.40 & 85.29 & 67.09 & 77.12  & 0.10 \\
ConvNeXt V2~\cite{CVPR2023ConvNeXtV2} & IN      & 105.00M & 40.71G & 56.38 & 81.36 & 63.59 & 71.48  & 9.09 & 57.30 & 85.46 & 66.85 & 74.86  & 2.00 \\
 LSKNet~\cite{ICCV2023LSKNet} & IN      & 30.98M & 22.52G & 58.13 & 81.31 & 63.61 & 78.51  & 9.09 & 58.74 & 86.86 & 67.05 & 78.72  & 2.34 \\
  \hline
  \multicolumn{14}{l}{\textit{End2end}}  \\
  \hline
  DETR~\cite{carion2020end} & IN     & 41.56M & 6.95G & 8.89 & 13.39 & 1.26 & 6.70 & 14.21 & 3.85 & 5.38 & 0.82 & 2.96 & 6.24 \\
  Deformable DETR~\cite{zhu2020deformable} & IN     & 40.10M & 14.62G & 65.02 & \underline{89.96} & 32.99 & 70.28 & \textbf{66.83} & 66.08 & \textbf{92.37} & 31.27 & 73.16 & 67.52 \\
  DAB-DETR~\cite{liu2022dab} & IN     & 43.70M & 10.49G & 43.88 & 80.25 & 3.47 & 52.20  & 39.61 & 44.17 & 84.20 & 1.58 & 52.36 & 38.56 \\
  Conditional DETR~\cite{meng2021conditional} & IN     & 43.35M & 9.79G & 42.26 & 78.63 & 9.8 & 54.13  & 26.47 & 40.01 & 82.86 & 1.21 & 54.34 & 21.64 \\
  \hline
  \rowcolor[rgb]{0.9,0.9,0.9}$\star$ \textbf{DenoDet (Ours)}  & IN & 34.23M & 12.89G  & \textbf{69.90}   &  89.40 & \textbf{47.87} & \underline{78.92}  & 63.41 & \textbf{71.21} & 90.74 & \textbf{46.07} & \textbf{79.60} & \textbf{68.41} \\
  \end{tabular}
  \vspace{-1\baselineskip}
  \end{table*}

  \begin{table*}[htbp]
    \renewcommand\arraystretch{1.2}
    \footnotesize
    \centering
      \vspace{-1\baselineskip}
    \caption{Comparison with SOTA methods on the \textbf{SAR-Aircraft-1.0} dataset.}
    \label{tab:aircraft}
    \vspace{-2pt}
    \setlength{\tabcolsep}{2pt}
    \begin{tabular}{l|c|c|accccccc|accccccc}
      \multirow{2}{*}{Method} & \multirow{2}{*}{Pre.}     & \multirow{2}{*}{FLOPs $\downarrow$} & \multicolumn{8}{c|}{Average Precision (AP'07)}  & \multicolumn{8}{c}{Average Precision (AP'12)} \\
        &  &   & \textbf{mAP} &  A220  & A320    &  A330   & ARJ21 & B737 & B787 & Other & \textbf{mAP} & A220  & A320    &  A330   & ARJ21 & B737 & B787 & Other \\
    \Xhline{1pt}
    \multicolumn{19}{l}{\textit{One-stage}}  \\
    \hline
  
    FCOS~\cite{tian2019fcos} & IN    & 51.58G & 62.63 & 62.27 & 93.19 & 63.65 & \textbf{63.10} & 40.06 & 52.37 & 63.79 & 63.76 & 64.63 & \underline{95.16} & 64.61 & 63.02 & 39.36 & 53.05 & 66.50 \\
  
    GFL~\cite{li2020generalized} & IN    & 32.27G & 66.90 & 59.90 & 87.34 & 92.88 & 62.53 & \underline{46.16} & \textbf{61.82} & 57.64 & \underline{68.44} & 61.19 & 90.61 & 94.75 & \textbf{66.10} & \textbf{45.98} & \textbf{62.11} & 58.38 \\
  
    RepPoints~\cite{yang2019reppoints} & IN    & 48.50G & \underline{67.13} & 64.27 & 89.75 & 86.28 & 61.28 & 41.50 & 59.73 & \underline{67.10} & 68.09 & 65.72 & 93.03 & 86.91 & 62.88 & 40.01 & 59.80 & 68.29 \\
  
    ATSS~\cite{zhang2020bridging} & IN    & 51.58G & 66.01 & 58.57 & 88.60 & \underline{96.57} & 60.14 & 37.79 & 59.18 & 61.25 & 66.71 & 59.84 & 92.03 & 96.74 & 61.78 & 36.12 & 59.10 & 61.36 \\
  
    CenterNet~\cite{zhou2019objects} & IN    & 51.57G & 64.11 & 58.03 & 93.03 & \textbf{97.02} & 61.90 & 36.41 & 41.74 & 60.61 & 64.90 & 58.15 & 96.58 & \underline{97.36} & 62.13 & 34.99 & 42.28 & 62.81 \\
  
    PAA~\cite{kim2020probabilistic} & IN    & 51.58G & 66.79 & \underline{66.26} & 89.56 & 96.41 & 62.20 & 35.90 & 54.10 & 63.12 & 67.56 &        \underline{67.56} & 93.51 & 97.17 & 62.91 & 34.69 & 53.04 & 64.07 \\
  
    PVT-T~\cite{wang2021pyramid} & IN    & 42.30G & 61.64 & 52.70 & 82.01 & 85.94 & 61.07 & 32.63 & 54.94 & 62.15 & 62.43 &        53.32 & 84.46 & 87.85 & 62.12 & 30.57 & 55.56 & 63.15 \\
  
    RetinaNet~\cite{lin2017focal} & IN    & 52.88G & 66.47 & 65.77 & 95.10 & 82.31 & 54.79 & 41.22 & \underline{60.59} & 65.52 & 67.26 & 66.66 & \textbf{97.09} & 84.08 & 54.37 & 39.33 & \underline{61.16} & 68.10 \\
  
    TOOD~\cite{feng2021tood} & IN    & 50.53G & 62.66 & 53.94 & 82.20 & 91.86 & 60.51 & 35.36 & 53.81 & 60.95 & 62.98 & 53.81 & 83.91 & 93.99 & 60.51 & 33.91 & 53.50 & 61.21 \\

DDOD~\cite{ACM2021DDOD} & IN    & 45.60G & 62.66 & 53.02 & 87.34 & 96.12 & 59.95 & 35.31 & 44.54 & 62.39 & 63.10 & 52.44 & 91.45 & 96.31 & 61.40 & 34.83 & 42.40 & 62.87 \\
  
    VFNet~\cite{zhang2021varifocalnet} & IN    & 48.39G & 66.17 & 60.94 & 90.91 & 95.25 & 61.29 & 39.41 & 56.46 & 58.93 & 66.90 &  62.00 & 95.57 & 96.14 & 60.76 & 36.63 & 56.76 & 60.43 \\
    
    YOLOF~\cite{chen2021you} & IN    & 26.33G & 66.25 & 60.24 & 90.75 & 88.81 & 61.00 & 36.30 & 59.64 & 66.99 & 67.71 & 60.97 & 94.85 & 91.31 & 63.37 & 34.35 & 60.42 & \underline{68.71} \\

YOLOX~\cite{Arixv2021YOLOX} & IN    & 8.53G & 63.65 & 63.93 & 89.61 & 97.54 & 52.01 & 24.19 & 56.63 & 61.67 & 65.05 & 64.91 & 91.27 & 97.88 & 55.51 & 22.88 & 60.65 & 62.26 \\
    
    AutoAssign~\cite{zhu2020autoassign} & IN    & 51.84G & 62.36 & 59.40 & 82.98 & 92.62 & 57.49 & 32.62 & 51.21 & 60.18 & 63.11 & 60.61 & 86.39 & 94.02 & 57.87 & 30.86 & 51.31 & 60.74 \\
    \hline
    \multicolumn{19}{l}{\textit{Two-stage}}  \\
    \hline
  
    Faster R-CNN~\cite{ren2015faster} & IN    & 63.21G & 64.71 & 56.96 & 88.09 & 96.99 & 52.78 & 35.74 & 59.08 & 63.32 & 65.11 & 58.63 & 90.53 & 97.47 & 56.82 & 32.31 & 57.07 & 62.94 \\
  
    Cascade R-CNN~\cite{cai2019cascade} & IN    & 91.00G & 64.87 & 51.87 & \underline{96.32} & \underline{97.51} & \underline{62.65} & 30.27 & 58.39 & 57.09 & 65.27 & 52.76 & 96.58 & 97.73 & \underline{65.66} & 27.21 & 56.10 & 60.88 \\
  
    Dynamic R-CNN~\cite{zhang2020dynamic} & IN    & 63.21G & 64.59 & 51.03 & 92.49 & \textbf{97.56} & 59.63 & 32.37 & 56.21 & 62.82 & 65.38 & 53.71 & 93.23 & \textbf{98.02} & 59.03 & 32.07 & 58.86 & 62.72 \\
  
    Grid R-CNN~\cite{lu2019grid} & IN    & 0.18T & 64.15 & 54.05 & \textbf{98.08} & 97.16 & 60.47 & 28.94 & 47.73 & 62.61 & 64.04 & 54.48 & 98.42 & 97.36 & 59.54 & 26.80 & 49.14 & 62.51 \\
  
    Libra R-CNN~\cite{pang2019libra} & IN    & 64.02G & 63.46 & 55.47 & 94.35 & 88.72 & 53.63 & 37.03 & 52.90 & 62.12 & 65.47 & 56.27 & 96.79 & 93.63 & 57.12 & 35.03 & 54.89 & 64.55 \\
     ConvNeXt~\cite{CVPR2022ConvNeXt} & IN    & 63.85G & 67.41 & 63.44 & 92.25 & 97.53 & 69.35 & 38.58 & 49.06 & 61.64 & 66.98 & 63.51 & 93.71 & 97.66 & 67.61 & 36.17 & 49.06 & 61.11 \\
    ConvNeXt V2~\cite{CVPR2023ConvNeXtV2} & IN    & 0.12T & 68.04 & 61.13 & 93.77 & 97.16 & 59.73 & 36.28 & 62.38 & 65.82 & 68.72 & 62.71 & 94.88 & 97.30 & 59.93 & 34.24 & 63.19 & 68.76 \\
    LSKNet~\cite{ICCV2023LSKNet} & IN    & 53.73G & 67.58 & 59.15 & 99.06 & 96.83 & 58.85 & 37.34 & 56.84 & 64.98 & 68.26 & 59.82 & 99.21 & 97.21 & 60.80 & 35.56 & 57.47 & 67.79 \\
    \hline
    \multicolumn{19}{l}{\textit{End2end}}  \\
    \hline
    DETR~\cite{carion2020end} & IN    & 24.94G & 10.61 & 19.55 & 0.78 & 0.47 & 4.15 & 9.11 & 20.95 & 19.25 & 8.27 & 15.80 & 0.56 & 0.45 & 2.45 & 6.14 & 18.97 & 13.55 \\
    Deformable DETR~\cite{zhu2020deformable} & IN    & 51.78G & 62.43 & \textbf{66.76} & 69.52 & 89.10 & 59.17 & 40.98 & 48.62 & 62.86 & 63.61 & \textbf{68.35} & 71.74 & 94.05 & 60.68 & 38.82 & 47.83 & 63.81 \\
    
    DAB-DETR~\cite{liu2022dab} & IN    & 28.94G & 53.62 & 59.82 & 88.39 & 18.84 & 62.26 & 34.52 & 54.34 & 57.15 & 54.16 & 60.80 & 91.10 & 18.97 & 63.61 & 33.20 & 54.04 & 57.42 \\
  
    Conditional DETR~\cite{meng2021conditional} & IN    & 28.09G & 62.25 & 53.74 & 89.38 & 78.83 & 61.75 & 37.44 & 61.45 & 53.19 & 63.02 & 54.27 & 93.30 & 80.47 & 62.07 & 35.92 & 61.99 & 53.14 \\
    \hline
    \rowcolor[rgb]{0.9,0.9,0.9}$\star$ \textbf{DenoDet (Ours)}  & IN    & 48.53G & \textbf{68.60}   &  64.82 & 90.54 & 93.82 & 59.58 & \textbf{46.32} & 56.04 & \textbf{69.05} & \textbf{69.56} & 66.53 & 93.43 & 94.84 & 60.80 & \underline{44.85} & 55.74 & \textbf{70.71} \\
    \end{tabular}
    \vspace{-1\baselineskip}
    \end{table*}

\begin{table*}[htbp]
  \renewcommand\arraystretch{1.2}
  \centering
  \caption{Comparison with SOTA methods on the \textbf{SARDet-100K} dataset.}
  \label{tab:sardet}
  \vspace{-2pt}
  \footnotesize
  \setlength{\tabcolsep}{2pt}{
  \begin{tabular}{l|c|cc|accccccccccc}
Method & \small Pre. & \small FLOPs $\downarrow$ & \small \#Params $\downarrow$ & \small\textbf{mAP} & \small AP@50  & \small AP@75  & $\small AP_{\small S}$  & $\small AP_{\small M}$  & $\small AP_{\small L}$ & \small Ship & \small Aircraft & \small Car & \small Tank & \small Bridge & \small Harbor\\
  \Xhline{1pt}
  \multicolumn{15}{l}{\textit{One-stage}}  \\
  \hline
FCOS~\cite{tian2019fcos}                  & IN       & 51.57G    & 32.13M                     & 52.52        & 85.82 & 54.93 & 47.01 & 66.13 & 57.82 & 59.79 & 55.44 & 60.75 & 41.78 & 34.17 & 63.44               \\
  
 GFL~\cite{li2020generalized}                & IN       & 52.36G  & 32.27M                       & \underline{55.01}         & 85.16 & 58.87 & 49.44 & 67.29 & \underline{60.45} & 63.92 & \textbf{57.63} & 62.29 & 44.80 & 36.41 & 65.04                 \\
  
RepPoints~\cite{yang2019reppoints}                & IN       & 48.49G  & 36.82M                       & 51.66         & 86.43 & 53.99 & 46.66 & 63.26 & 53.78 & 60.85 & 55.50 & 61.13 & 40.69 & 35.12 & 56.71                 \\
  
ATSS~\cite{zhang2020bridging}                & IN       & 51.57G   & 32.13M                     & 54.95        & 87.60 & 58.25 & 49.89 & 67.94 & 58.97 & 61.53 & 55.94 & 61.77 & \underline{46.20} & 37.22 & \textbf{67.48}               \\
  
CenterNet~\cite{zhou2019objects}                & IN       & 51.55G   & 32.12M                      & 53.91         & 86.17 & 57.31 & 48.88 & 66.22 & 57.74 & 61.24 & 56.35 & 61.74 & 45.31 & 35.91 & 63.29                 \\

PAA~\cite{kim2020probabilistic}                & IN       & 51.57G  & 32.13M                       & 52.20         & 85.71 & 54.80 & 46.00 & 63.90 & 57.61 & 60.16 & 56.17 & 60.09 & 41.07 & 35.96 & 60.12               \\
  
 PVT-T~\cite{wang2021pyramid}                & IN       & 42.19G   & 21.43M                      & 46.10         & 77.55 & 49.00 & 38.01 & 59.53 & 53.35 & 53.30 & 52.91 & 59.03 & 30.20 & 22.51 & 59.11                 \\
  
RetinaNet~\cite{lin2017focal}                & IN       & 52.77G & 36.43M                       & 46.48         & 77.74 & 48.94 & 40.25 & 59.35 & 50.26 & 55.36 & 54.00 & 60.88 & 32.72 & 24.81 & 51.12                 \\
  
TOOD~\cite{feng2021tood}                & IN       & 50.52G & 30.03M                       & 54.65         & 86.88 & 58.41 & \underline{50.20} & \underline{66.72} & 58.60 & 62.28 & 55.61 & 62.53 & 45.96 & 36.64 & 65.24                 \\

 DDOD~\cite{ACM2021DDOD}                & IN       & 45.58G & 32.21M                       & 54.02         & 86.64 & 57.23 & 49.33 & 64.70 & 58.02 & 62.39 & 56.08 & 62.48 & 43.98 & 36.34 & 62.87                 \\
  
VFNet~\cite{zhang2021varifocalnet}                & IN       & 48.38G & 32.72M                        & 53.01         & 84.32 & 56.32 & 47.37 & 65.39 & 57.99 & 62.14 & 55.84 & 61.97 & 42.08 & 34.11 & 62.28                 \\
  
AutoAssign~\cite{zhu2020autoassign}                & IN       & 51.83G  & 36.26M                       & 53.95        & \textbf{89.58} & 55.96 & 50.14 & 63.40 & 54.73 & 62.03 & 55.70 & 61.69 & \textbf{48.55} & \underline{38.25} & 57.45                 \\

YOLOF~\cite{chen2021you}                & IN       & 26.32G & 42.46M                        & 42.83        & 74.95 & 43.18 & 33.73 & 56.19 & 53.57 & 52.62 & 52.64 & 52.71 & 22.86 & 23.74 & 52.42                 \\

YOLOX~\cite{Arixv2021YOLOX}                & IN       & 8.53G  & 8.94M                       & 34.08        & 66.77 & 31.31 & 28.49 & 43.06 & 28.95 & 46.08 & 46.83 & 53.43 & 26.26 & 13.14 & 18.95                 \\
  
  \hline
  \multicolumn{5}{l}{\textit{Two-stage}}  \\
  
  \hline
Faster R-CNN~\cite{ren2015faster}                & IN       & 63.2G  & 41.37M                       & 39.22         & 70.04 & 39.87 & 32.55 & 47.23 & 42.02 & 50.45 & 50.36 & 57.82 & 24.90 & 18.69 & 33.11                 \\

Cascade R-CNN~\cite{cai2019cascade}                & IN       & 90.99G     & 69.17M                    & 53.55         & 87.33 & 56.81 & 49.09 & 62.89 & 48.68 & \textbf{66.99} & 56.43 & \underline{63.25} & 44.35 & 36.89 & 53.81                \\

Dynamic R-CNN~\cite{zhang2020dynamic}                & IN       & 63.2G      & 41.37M                   & 49.75         & 80.96 & 53.91 & 43.12 & 59.72 & 54.77 & 61.32 & 53.86 & 60.00 & 33.68 & 34.40 & 55.25                 \\

 Grid R-CNN~\cite{lu2019grid}                & IN       & 0.18T            & 64.47M             & 50.05         & 80.58 & 53.49 & 42.43 & 62.01 & 52.70 & 60.43 & 55.61 & 61.94 & 36.03 & 31.16 & 55.13                \\
  
Libra R-CNN~\cite{pang2019libra}                & IN       & 64.02G       & 41.64M                  & 52.09         & 83.54 & 55.81 & 45.85 & 63.52 & 55.40 & 61.32 & 54.03 & 61.56 & 38.12 & 35.97 & 61.50                 \\

ConvNeXt~\cite{CVPR2022ConvNeXt}                & IN       & 63.84G          & 45.07M              & 53.15         & 85.52 & 57.28 & 45.67 & 64.55 & 58.61 & 60.55 & 57.35 & 62.13 & 38.12 & 36.81 & 63.95                 \\

ConvNeXtV2~\cite{CVPR2023ConvNeXtV2}                & IN       & 0.12T         & 0.11G                & 53.91         & 86.01 & \underline{58.90} & 47.63 & 64.67 & 59.57 & 61.48 & 55.83 & 63.23 & 39.65 & \textbf{39.16} & 64.09                 \\

LSKNet~\cite{ICCV2023LSKNet}                & IN       & 53.73G            & 30.99M             & 52.39         & 85.07 & 56.96 & 45.15 & 63.59 & 59.16 & 59.33 & 56.76 & 62.74 & 36.09 & 35.01 & 64.38                 \\

  \hline
  \multicolumn{5}{l}{\textit{End2end}}  \\
  
  \hline
DETR~\cite{carion2020end}                & IN       & 24.94G              & 41.56M           & 45.73         & 78.57 & 46.87 & 37.01 & 58.16 & 55.58 & 54.94 & 51.17 & 50.11 & 26.06 & 32.80 & 59.31                 \\

Deformable DETR~\cite{zhu2020deformable}                & IN       & 51.78G & 40.10M                        & 52.00         & \underline{88.77} & 54.03 & 46.99 & 63.58 & 58.55 & 60.94 & 54.16 & 61.22 & 39.14 & 36.09 & 60.46               \\
  
DAB-DETR~\cite{liu2022dab}                & IN       & 28.94G             & 43.70M            & 43.31         & 78.14 & 43.10 & 34.82 & 56.34 & 52.62 & 53.16 & 50.32 & 49.47 & 24.06 & 28.47 & 55.07                 \\

Conditional DETR~\cite{meng2021conditional}                & IN       & 28.09G & 43.45M                         & 44.04         & 77.88 & 44.40 & 35.25 & 56.47 & 52.86 & 52.77 & 49.58 & 51.00 & 22.73 & 29.98 & 58.16                 \\
  
  \hline
\rowcolor[rgb]{0.9,0.9,0.9}$\star$ \textbf{DenoDet (Ours)}    & IN     & 52.69G & 65.78M                            & \textbf{55.88}   & 85.81 & \textbf{60.16} & \textbf{50.63} & \textbf{68.47} & \textbf{60.96} & \underline{64.91} & \underline{57.36} & \textbf{63.66} & 45.79 & 36.39 & \underline{67.17}   \\
  \end{tabular}}
  \vspace{\baselineskip}
  \end{table*}

\subsubsection{\textbf{Results on MSAR.}}

As presented in Table~\ref{tab:msar}, our proposed detection framework, DenoDet, was compared with 23 state-of-the-art methods on the challenging MSAR dataset. DenoDet excelled, achieving a mean Average Precision (mAP) of \textbf{69.90\%} on the AP'07 metric and \textbf{71.21\%} on AP'12. Notably, DenoDet's enhanced performance on the MSAR dataset does not result in a proportional increase in computational complexity. It offers a 2.5\% improvement in detection accuracy over its baseline detector, FCOS, with only a slight increase in parameters and virtually unchanged FLOPs. This balance of high efficiency and effectiveness highlights the practicality of DenoDet for real-world SAR target detection, where both precision and computational resource management are critical.

\subsubsection{\textbf{Results on SAR-Aircraft-1.0.}}

In the SAR-Aircraft-1.0 dataset evaluation, described in Table~\ref{tab:aircraft}, DenoDet outperformed 25 other methods, securing the leading position with an mAP of \textbf{68.60\%}, according to the PASCAL VOC 2007~\cite{IJCV10VOC} benchmark. This achievement emphasizes DenoDet's role as a benchmark model for future SAR-based reconnaissance and surveillance research and operations.


\subsubsection{\textbf{Results on SARDet-100K.}}

Table~\ref{tab:sardet} presents the quantitative results of our DenoDet method compared to state-of-the-art approaches on the SARDet-100K dataset.
It is evident from the table that DenoDet achieves the highest performance, surpassing the second-best method, GFL, by a significant margin of 0.87\% in terms of mAP.
Since SARDet-100K dataset is currently the largest public SAR target detection dataset, this substantial improvement demonstrates the superior detection accuracy and robustness of our approach.

\subsection{Ablation Study} \label{subsec:ablation}

\subsubsection{\textbf{The Necessity of Frequency Transformation: Frequency Domain vs Spatial Domain.}}

Within the scope of this study, our objective was to validate the hypothesis that dynamic soft-thresholding in the frequency domain significantly enhances target detection in SAR imagery compared to the spatial domain. Our experimental setup, outlined in Table~\ref{tab:frequency}, divided the analysis into groups based on the use of Discrete Cosine Transform (DCT)/Inverse Discrete Cosine Transform (IDCT) and our dynamic soft-thresholding approach. The results were revealing. Implementing our Attention as Dynamic Soft-Thresholding technique without DCT/IDCT increased performance by 0.36\% on the MSAR benchmark. However, integrating this technique with DCT/IDCT resulted in a substantial 1.42\% performance improvement, achieved without increasing Floating Point Operations (FLOPs). This significant improvement substantiates our claim that frequency domain transformations, due to their global semantic aggregation, enhance the efficacy of attentional soft-thresholding compared to operations in the spatial domain.

Additionally, our experiments demonstrated that DCT/IDCT alone, without attention mechanisms, decreased the model's performance from 67.52\% to 67.00\%, highlighting the effectiveness of frequency domain attention. This decline indicates that our attention mechanism can offset the precision loss associated with DCT/IDCT. While increasing the numerical precision using 64-bit floating-point computations improved performance to 67.37\%, it also introduced greater computational demand and slower inference. Consequently, we maintained 32-bit precision for an optimal balance of performance and computational efficiency.

\setlength{\tabcolsep}{4pt}
\begin{table*}[htbp]
\centering
\vspace{-1\baselineskip}
\caption{Ablation study on the necessity of frequency transformation: frequency domain vs spatial domain}
\label{tab:frequency}
\vspace{-.5\baselineskip}
\small
\begin{tabular}{Sc Sc Sc Sc Sc Sc Sc Sc Sc Sc Sc Sc}
\toprule
\multirow{2}{*}{DCT/IDCT} & \multirow{2}{*}{Attention} & \multirow{2}{*}{Float} & \multicolumn{3}{c}{MSAR} & \multicolumn{3}{c}{SAR-AIRcraft-1.0} & \multicolumn{3}{c}{AIR-SARShip-1.0} \\
\cmidrule(lr){4-6} \cmidrule(lr){7-9} \cmidrule(lr){10-12}
&  &  & FPS & mAP (07) & mAP (12) & FPS & mAP (07) & mAP (12) & FPS & mAP (07) & mAP (12) \\
\midrule
\xmark & \xmark & 32 & 230.4 & 67.52 & 69.85 & 167.6 & 62.63 & 63.76 & 166.3 & 63.66 & 64.25\\
\xmark & $\checkmark$ & 32 & 230.1 & 67.88 & 70.09 & 167.4 & 64.01 & 60.40 & 168.4 & 63.70 & 63.50 \\
$\checkmark$ & \xmark & 32 & 225.0 & 67.00 & 69.26 & 167.3 & 60.72 & 61.49 & 167.4 & 61.91 & 62.63 \\
$\checkmark$ & \xmark & 64 & 220.3 & 67.37 & 69.57 & 152.3 & 61.58 & 62.05 & 156.3 & 62.06 & 62.05 \\
\rowcolor[rgb]{0.9,0.9,0.9} $\checkmark$ & $\checkmark$ & 32 & 215.0 & \textbf{68.94} & \textbf{70.46} & 166.4 & \textbf{64.18} & \textbf{65.17} & 164.9 & \textbf{65.91} & \textbf{66.70} \\
\bottomrule
\end{tabular}
\end{table*}

\setlength{\tabcolsep}{4pt}
\begin{table*}[htbp]
\centering
\caption{Ablation study on the necessity of space-channel transposition: spatial attention vs channel attention}
\label{tab:transposition}
\vspace{-.5\baselineskip}
\small
\begin{tabular}{Sc Sc Sc Sc Sc Sc Sc Sc}
\toprule
\multirow{2}{*}{DCT/IDCT} & \multirow{2}{*}{Space-Channel Transposition} & \multicolumn{2}{c}{MSAR} & \multicolumn{2}{c}{SAR-AIRcraft-1.0} & \multicolumn{2}{c}{AIR-SARShip-1.0} \\
\cmidrule(lr){3-4} \cmidrule(lr){5-6} \cmidrule(lr){7-8}
&  & mAP (07) & mAP (12) & mAP (07) & mAP (12) & mAP (07) & mAP (12) \\
\midrule

\xmark & \xmark & 67.52 & 69.85 & 62.63 & 63.76 & 63.66 & 64.25 \\


$\checkmark$ & \xmark & 68.12 & 69.81 & 64.20 & 64.84 & 60.30 & 61.47 \\

\rowcolor[rgb]{0.9,0.9,0.9} $\checkmark$ & $\checkmark$  & \textbf{68.94} & \textbf{70.46} & \textbf{64.18} & \textbf{65.17} & \textbf{65.91} & \textbf{66.70} \\
\bottomrule
\end{tabular}
\end{table*}

\setlength{\tabcolsep}{6pt}
\begin{table*}[htbp]
\centering
\caption{Ablation study on the necessity of deformable group FC: impact of groups of FC layers}
\label{tab:deformable}
\vspace{-.5\baselineskip}
\small
\begin{tabular}{Sc Sc Sc Sc Sc Sc Sc Sc}
\toprule
\multicolumn{2}{c}{\multirow{2}{*}{Group Number}} & \multicolumn{2}{c}{MSAR} & \multicolumn{2}{c}{SAR-AIRcraft-1.0} & \multicolumn{2}{c}{AIR-SARShip-1.0} \\
\cmidrule(lr){3-4} \cmidrule(lr){5-6} \cmidrule(lr){7-8}
&  & mAP (07) & mAP (12) & mAP (07) & mAP (12) & mAP (07) & mAP (12) \\
\midrule
\multirow{6}{*}{Fixed} & 1 & 67.47 & 69.83 & 62.65 & 63.20 & 64.68 & 64.76\\
  & 2 & 68.43 & 70.28 & \underline{66.10} & 66.94 & 64.28  & 64.59 \\
  & 4 & \underline{69.49} & \underline{70.81} & 65.26 & 66.28 & 65.26 & 66.27 \\
  & 8 & 68.36 & 70.38 & 66.09 & \underline{67.22} & \underline{66.13} & 66.69 \\
  & 16 & 69.04 & 70.71 & 64.09 & 65.25 & 65.41 & 66.63 \\
  & 32 & 68.51 & 70.42 & 63.41 & 64.04 & 65.71  & \underline{67.40} \\
\rowcolor[rgb]{0.9,0.9,0.9} \multicolumn{2}{l}{$\star$ \textbf{Deformable (ours)}} & \textbf{69.90} & \textbf{71.21} & \textbf{67.39} & \textbf{67.43} & \textbf{66.98} & \textbf{68.73} \\
\bottomrule
\end{tabular}
\end{table*}

\subsubsection{\textbf{The Necessity of Space-Channel Transposition: Spatial Attention vs Channel Attention.}}

In this research, we aimed to prove that in SAR image target detection, the capture of spatial dependencies is more critical than channel-wise relationships. The results, presented in Table~\ref{tab:transposition}, reveal that models emphasizing spatial relationships outperform those focusing on channel dependencies. Specifically, shifting our attention mechanism from channel to spatial orientation for the MSAR dataset resulted in a notable 0.77\% increase in performance. This outcome not only corroborates but also emphasizes the importance of transpose operations in our TransDeno module.

It is important to note that our TransDeno module, designed to exploit long-range spatial dependencies, differs from traditional spatial attention methods such as the 2D Convolution used by GENet and LSKNet. We employ fully connected layers, as commonly used in channel attention, with the critical addition of space-channel transposition to effectively capture spatial relationships within a fully connected framework.

\subsubsection{\textbf{The Necessity of Deformable Group FC: Impact of Groups of FC Layers}}

Lastly, we contend that our Deformable Group Fully Connected (DeGroFC) layer surpasses the conventional Group Fully Connected layer with static group configurations in SAR imagery target detection. The DeGroFC layer dynamically adjusts group numbers based on the SAR imagery content, demonstrating a flexible recognition of spatial interdependencies. The supporting evidence, detailed in Table~\ref{tab:deformable}, shows that there is no universally optimal group number for the static model; rather, it varies with the image characteristics. For the MSAR dataset, the DeGroFC layer improved performance by 0.4\% over the optimal static group configuration, confirming the advantage of our adaptive grouping approach.

\section{Analysis} \label{sec:analysis}

\subsection{\textbf{Feature Denoising Visualization}}



\begin{figure*}[htbp]
  \centering
  \vspace{-1\baselineskip}
  \includegraphics[width=.99\textwidth]{"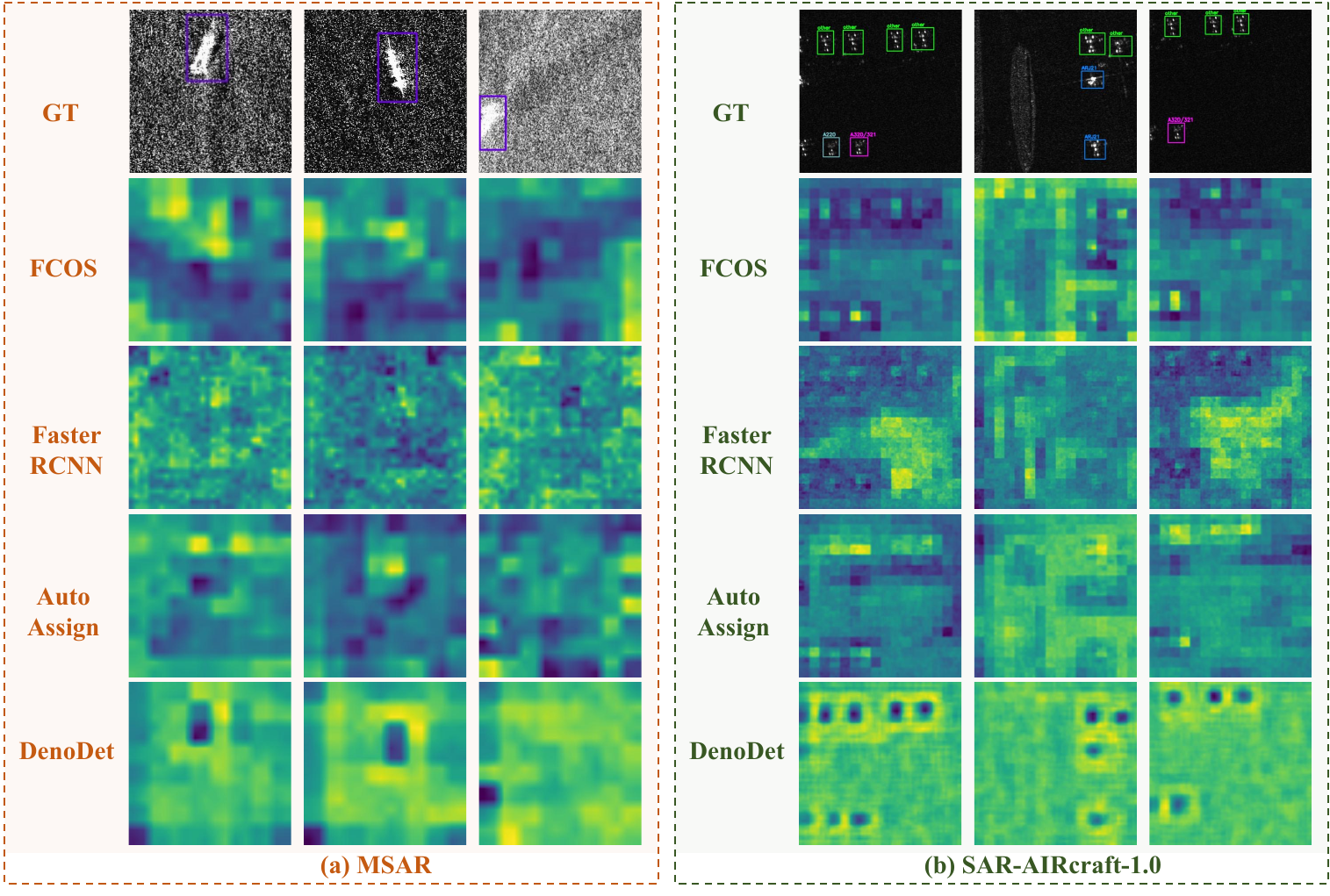"}
  \vspace{-.5\baselineskip}
  \caption{Visualization of noise suppression in feature maps by the TransDeno module: (a) MSAR, (b) SAR-Aircraft-1.0.
  }
  \label{fig:denoising}
\end{figure*}

In this part, we provide visual evidence supporting the central hypothesis of our research, utilizing Grad-CAM-based visualizations \cite{IJCV20GradCAM} to showcase the effective dynamic soft-thresholding denoising performance of our TransDeno module. Fig.~\ref{fig:denoising} offers a clear demonstration of this functionality.

Considering the MSAR dataset, infamous for its severe coherent noise, we observed the feature maps at the FPN's base level. While FCOS, the host detector of DenoDet, retains significant speckle noise, our DenoDet's feature maps are noticeably clear of such disturbances.


With the SAR-Aircraft-1.0 dataset, despite less pronounced coherent noise, it includes many background elements with high reflectivity. These elements act as noise, as they are not the primary targets and cause reflective interference. As shown in Fig.~\ref{fig:denoising} (b), DenoDet more effectively suppresses the features of these background elements, thus better highlighting the targets of interest than FCOS. These visualizations demonstrate that DenoDet not only improves detection accuracy, as evidenced by metrics like mean Average Precision (mAP), but also effectively reduces noise within the feature maps.

\begin{figure*}[htbp]
  \centering
  \includegraphics[width=.99\textwidth]{"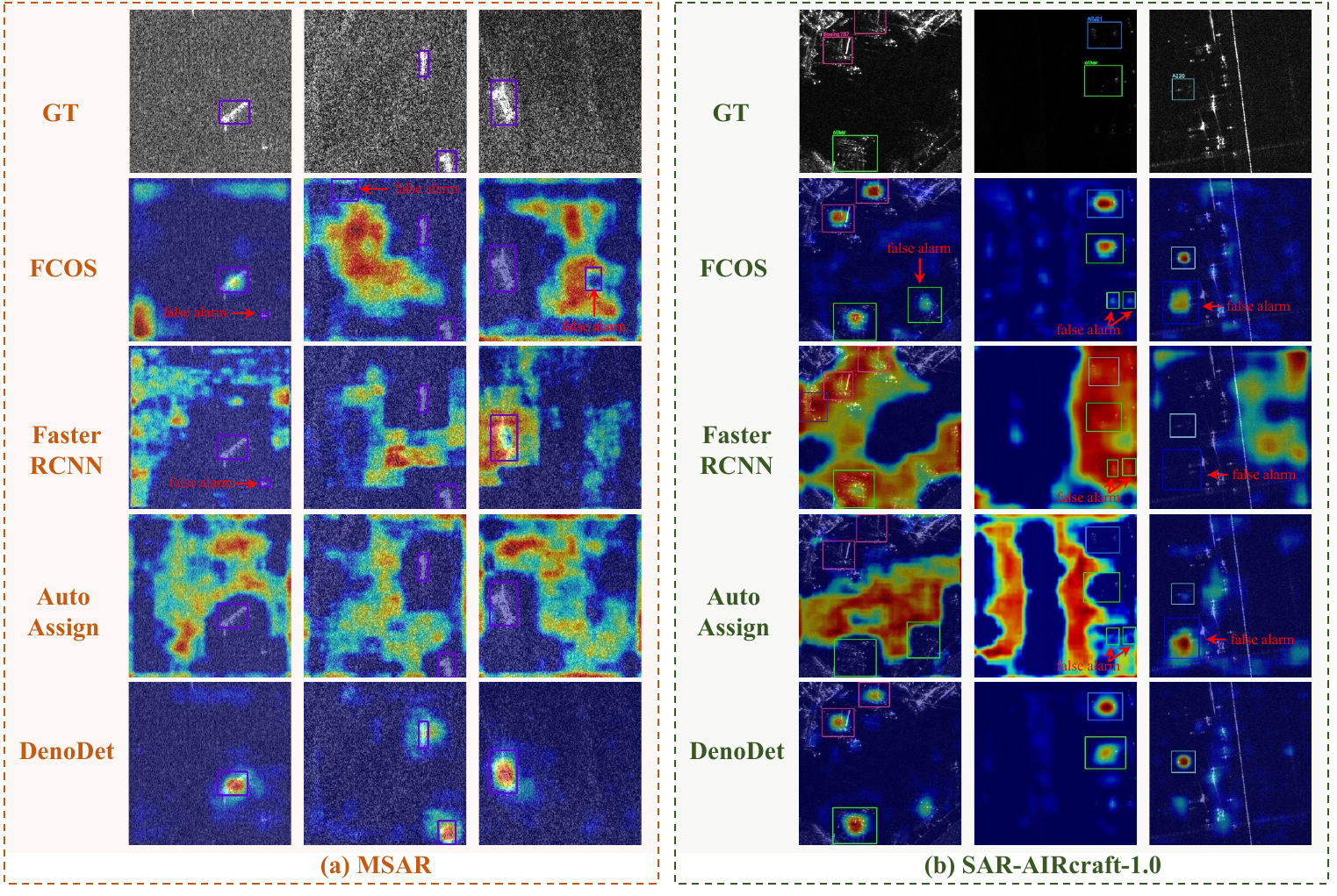"}
  \vspace{-.5\baselineskip}
  \caption{Comparative visualization of false alarm instances in SAR imagery. 
  }
  \label{fig:false-alarm}
\end{figure*}

\begin{figure*}[htbp]
  \centering
  \vspace{-1\baselineskip}
  \includegraphics[width=.99\textwidth]{"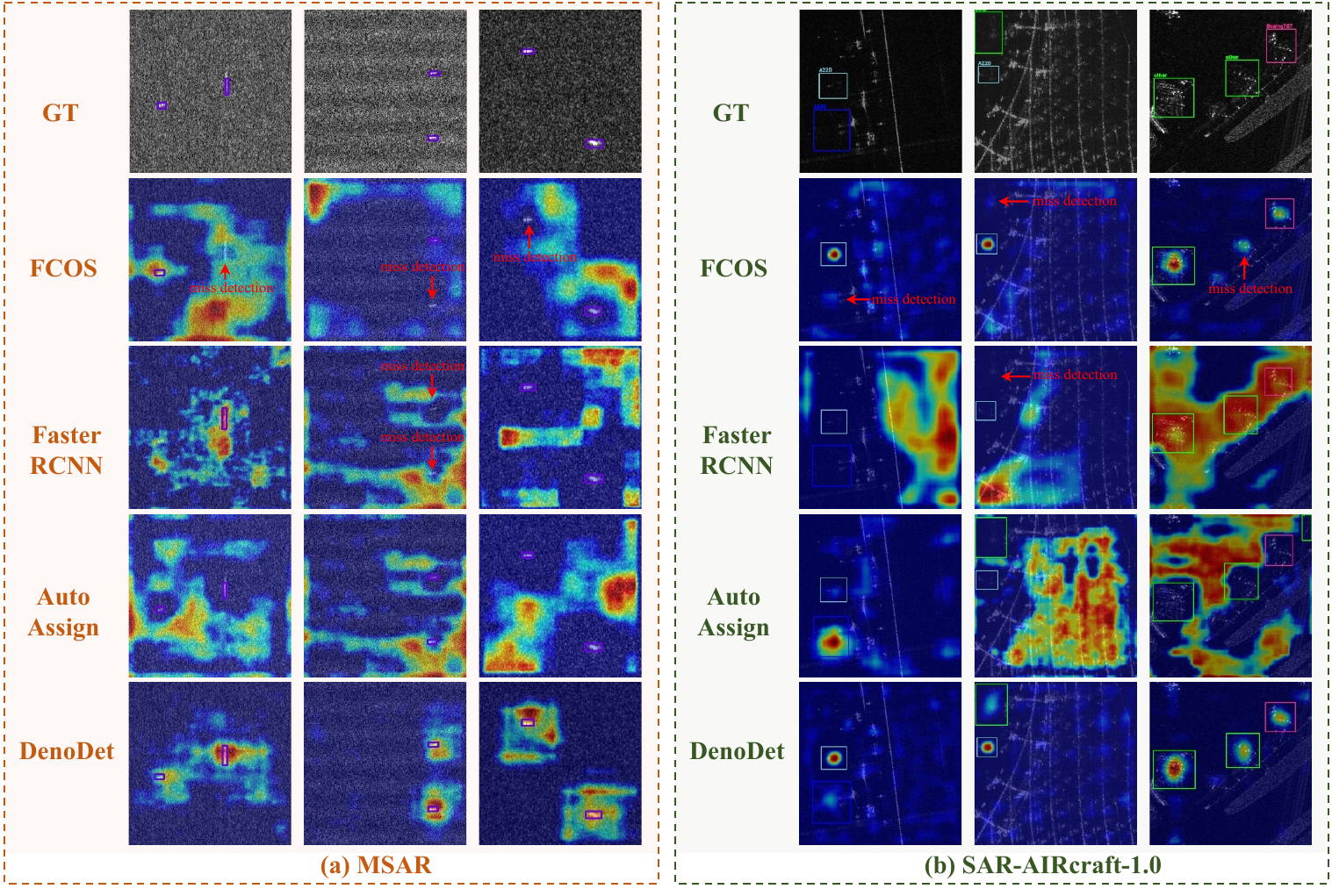"}
  \vspace{-.5\baselineskip}
  \caption{Detection capability comparison. Each subfigure illustrates instances of target miss detection.
  }
  \label{fig:miss-detection}
\end{figure*}

\begin{figure*}[htbp]
  \centering
  \includegraphics[width=.99\textwidth]{"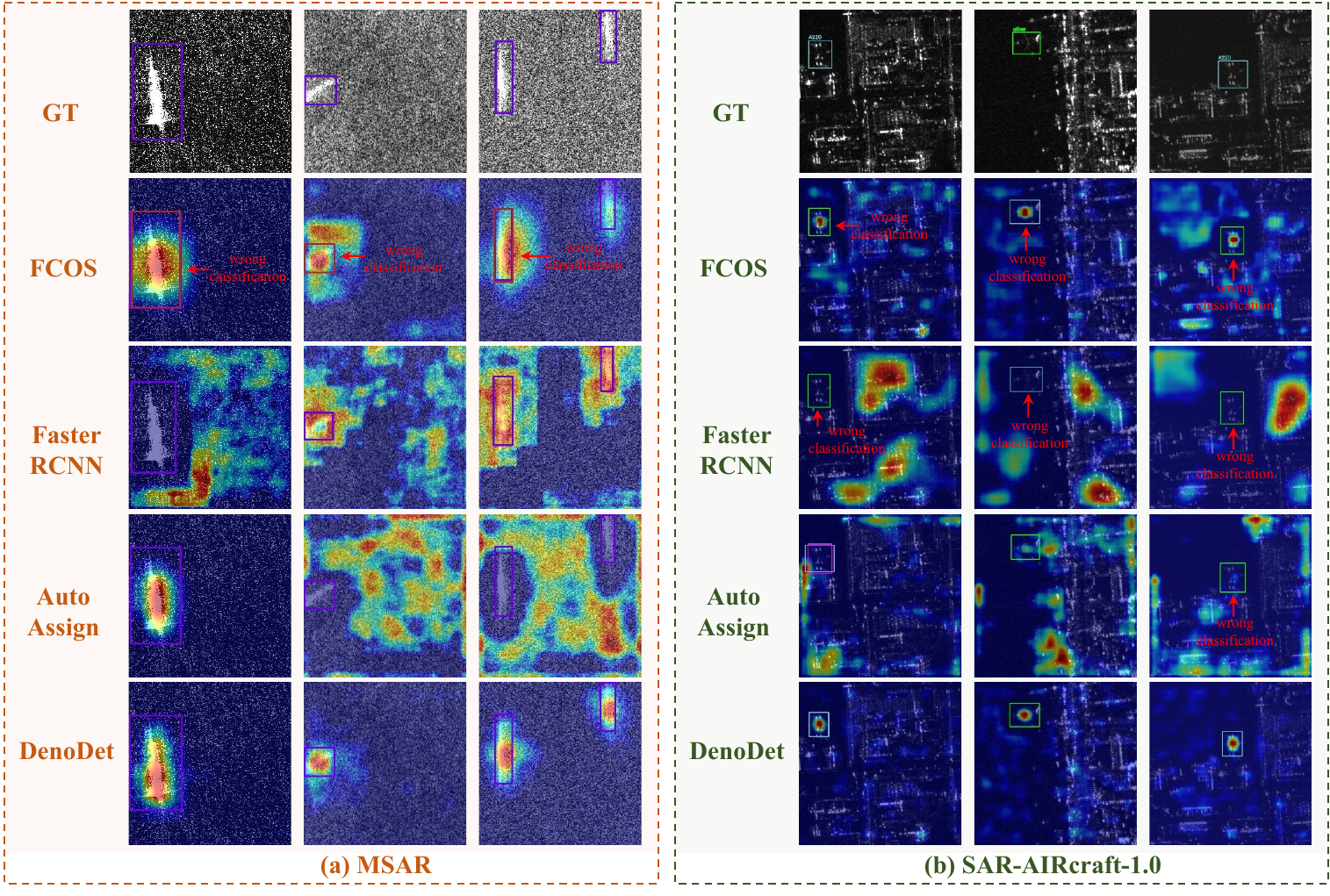"}
  \vspace{-.5\baselineskip}
  \caption{Discriminative efficacy comparison in SAR target classification. Each subfigure illustrates instances of incorrect target classification.
  }
  \label{fig:wrong-classification}
\end{figure*}

\subsection{\textbf{Detection Result Visualization}}

\subsubsection{Visualization on False Alarm Cases}

We further illustrate DenoDet's superior noise suppression and false alarm reduction in contrast to the baseline FCOS detector, as depicted in Fig.~\ref{fig:false-alarm}. Employing Grad-CAM visualization, we highlight DenoDet's efficacy.


In the challenging MSAR dataset, FCOS is prone to false positives due to highly reflective background regions that are not sufficiently suppressed during feature extraction. This often leads to misclassifications, particularly of the smallest target category—--Ship. Fig.~\ref{fig:false-alarm} (a) compares the effectiveness of DenoDet in reducing false positives by dampening background noise.


For the SAR-Aircraft-1.0 dataset, where target features are subtle, DenoDet outperforms FCOS in reducing false alarms, as shown in Fig.~\ref{fig:false-alarm} (b). The inclusion of the TransDeno module, with its DCT operation to capture global dependencies, extends the receptive field and significantly reduces false positives. By incorporating global context, DenoDet refines the feature space and enhances target detection fidelity.


\subsubsection{Visualization on Miss Detection Cases}

We also compare Grad-CAM-based feature maps to highlight DenoDet's reduced miss detection rates against the FCOS detector, as evidenced in Fig.~\ref{fig:miss-detection}. The TransDeno module's frequency domain transformations help to emphasize target details, especially for small-scale targets, and reduce miss detection rates.

\subsubsection{Visualization on Wrong Classification Cases}

Finally, we present a visual analysis using Grad-CAM-derived feature maps to illustrate DenoDet's enhanced classification capability, as evident in Fig.~\ref{fig:wrong-classification}. DenoDet significantly lowers misclassification rates across various target categories within complex datasets.

In the MSAR dataset, FCOS often confuses Bridge with Ship due to their spectral similarities. Similarly, in the SAR-Aircraft-1.0 dataset, it misclassifies A220 as Other. The TransDeno module counters this by using DCT for detailed frequency-domain feature extraction and a DeGroFC Layer that divides the spectral domain into adaptive subspaces for refined feature interaction, surpassing conventional approaches in preserving critical target details.





\section{Conclusion} \label{sec:conclusion}

In this work, we have addressed the persistent challenges of SAR target detection, specifically the issues of speckle noise and the detection of small, ambiguous targets. Our proposed DenoDet network, which incorporates an explicit frequency domain transform, has proven effective in recalibrating convolutional biases towards low-frequency components and facilitating a multi-subspace representation for improved target detection.
The introduction of our dynamic frequency domain attention module, TransDeno, has enabled a transform domain soft thresholding process that dynamically denoises the feature maps of the SAR imagery. This process effectively preserves important target signals while suppressing noise, enhancing the detection reliability in complex terrains. Moreover, the adaptability of our model is further improved by the deformable group fully-connected layer, which tailors the processing of frequency subspaces to the characteristics of the input features.
Our experiments have demonstrated that DenoDet achieves superior performance on various SAR target detection benchmarks. The comprehensive analysis validates the effectiveness and significance of our proposed model.


\bibliographystyle{IEEEtran}
\bibliography{./reference.bib}


\begin{IEEEbiography}[{\includegraphics[width=1.1in,height=1.5in,clip,keepaspectratio]{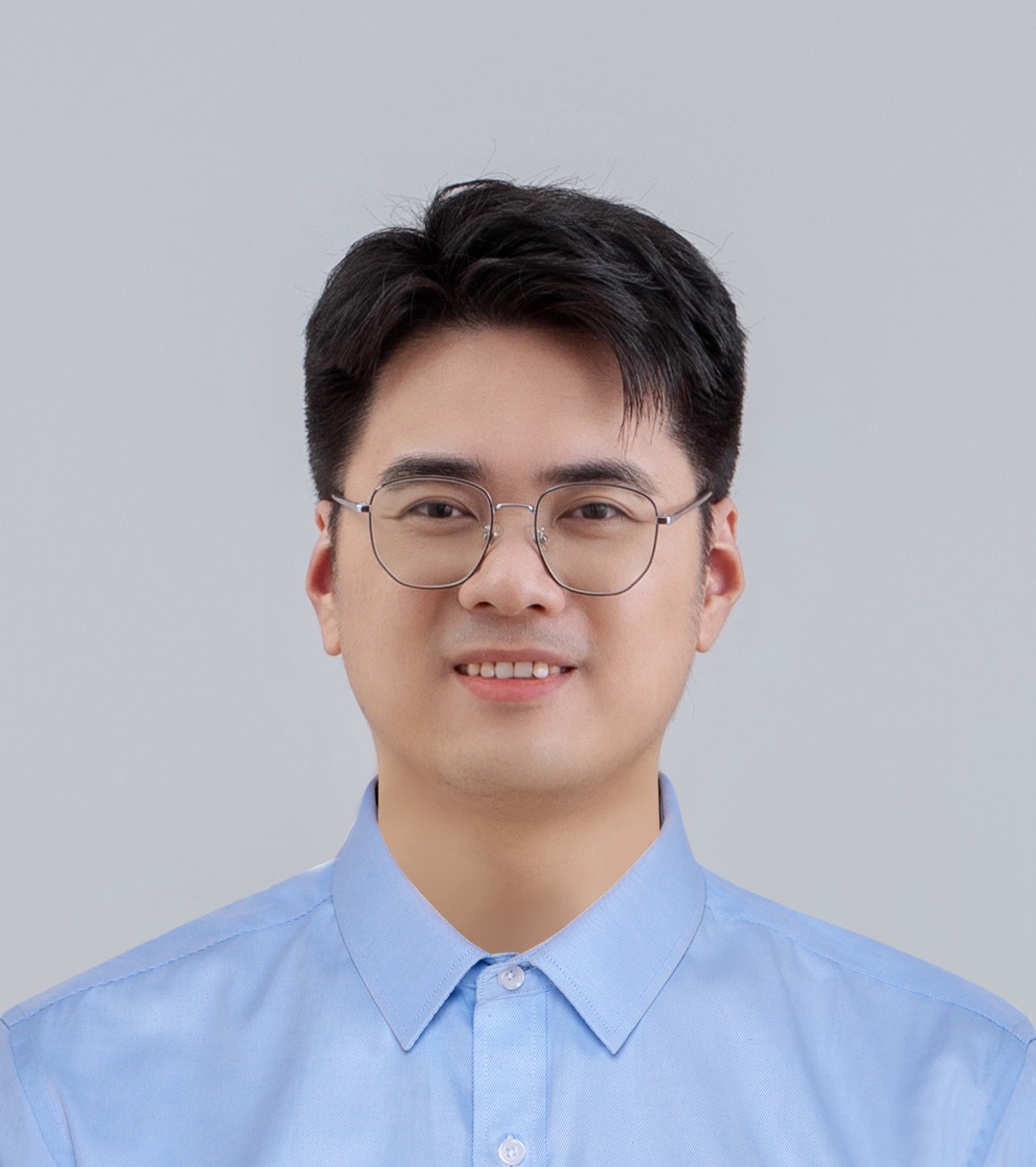}}]{Yimian Dai} received the Ph.D. degree in signal and information processing from Nanjing University of Aeronautics and Astronautics, Nanjing, China, in 2021, during which he honed his research skills as a visiting Ph.D. student at the University of Copenhagen and the University of Arizona between March 2018 and October 2020. Since 2021, he has been with the School of Computer Science and Engineering, Nanjing University of Science and Technology (NJUST), Nanjing, where he is currently an Assistant Researcher. His research interests include remote sensing, computer vision, and deep learning, with a focus on developing algorithms for object detection, data assimilation, and computational imaging to tackle real-world challenges. He has authored more than 20 peer-reviewed journal and conference papers such as IJCV, TGRS, TAES, etc.
\end{IEEEbiography}

\begin{IEEEbiography}[{\includegraphics[width=1.1in,height=1.35in,clip,keepaspectratio]{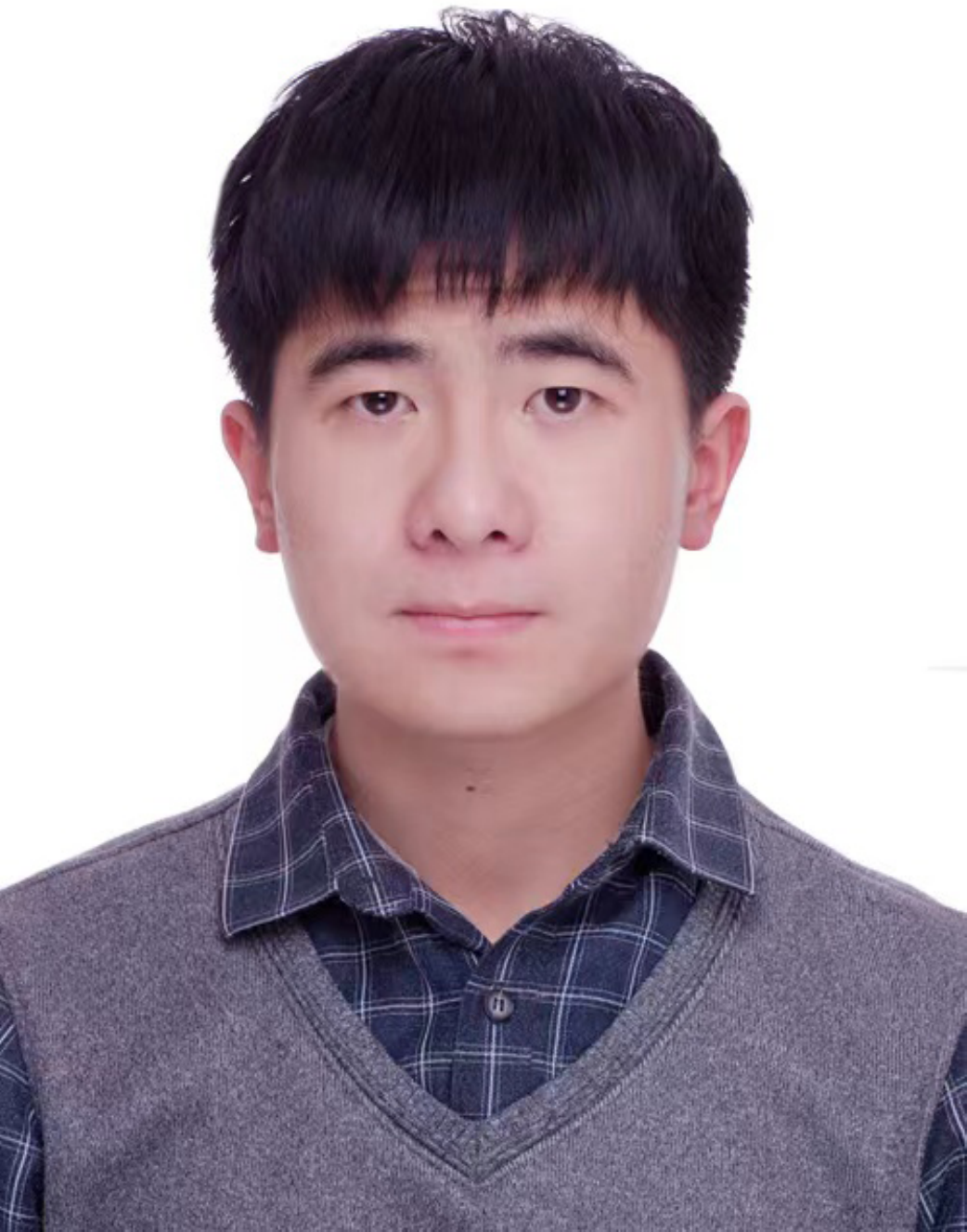}}]{Minrui Zou} received the B.E. degree from the Civil Aviation University of China, Tianjin, China, in 2022. And he is pursuing the M.S. degree in the School of Computer Science, Nanjing University of Posts and Telecommunications, Jiangsu, China. His research interests include machine learning and computer vision.
\end{IEEEbiography}

\begin{IEEEbiography}[{\includegraphics[width=1.1in,height=1.35in,clip,keepaspectratio]{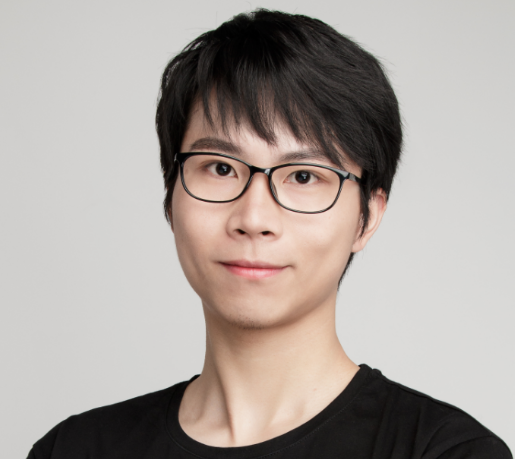}}]{Xiang Li} is an Associate Professor in College of Computer Science, Nankai University. He received the PhD degree from the Department of Computer Science and Technology, Nanjing University of Science and Technology (NJUST) in 2020. There, he started the postdoctoral career in NJUST as a candidate for the 2020 Postdoctoral Innovative Talent Program. In 2016, he spent 8 months as a research intern in Microsoft Research Asia, supervised by Prof. Tao Qin and Prof. Tie-Yan Liu. He was a visiting scholar at Momenta, mainly focusing on monocular perception algorithm. His recent works are mainly on: neural architecture design, CNN/Transformer, object detection/recognition, unsupervised learning, and knowledge distillation. He has published 20+ papers in top journals and conferences such as TPAMI, CVPR, NeurIPS, etc.
\end{IEEEbiography}

\begin{IEEEbiography}[{\includegraphics[width=1.1in,height=1.35in,clip,keepaspectratio]{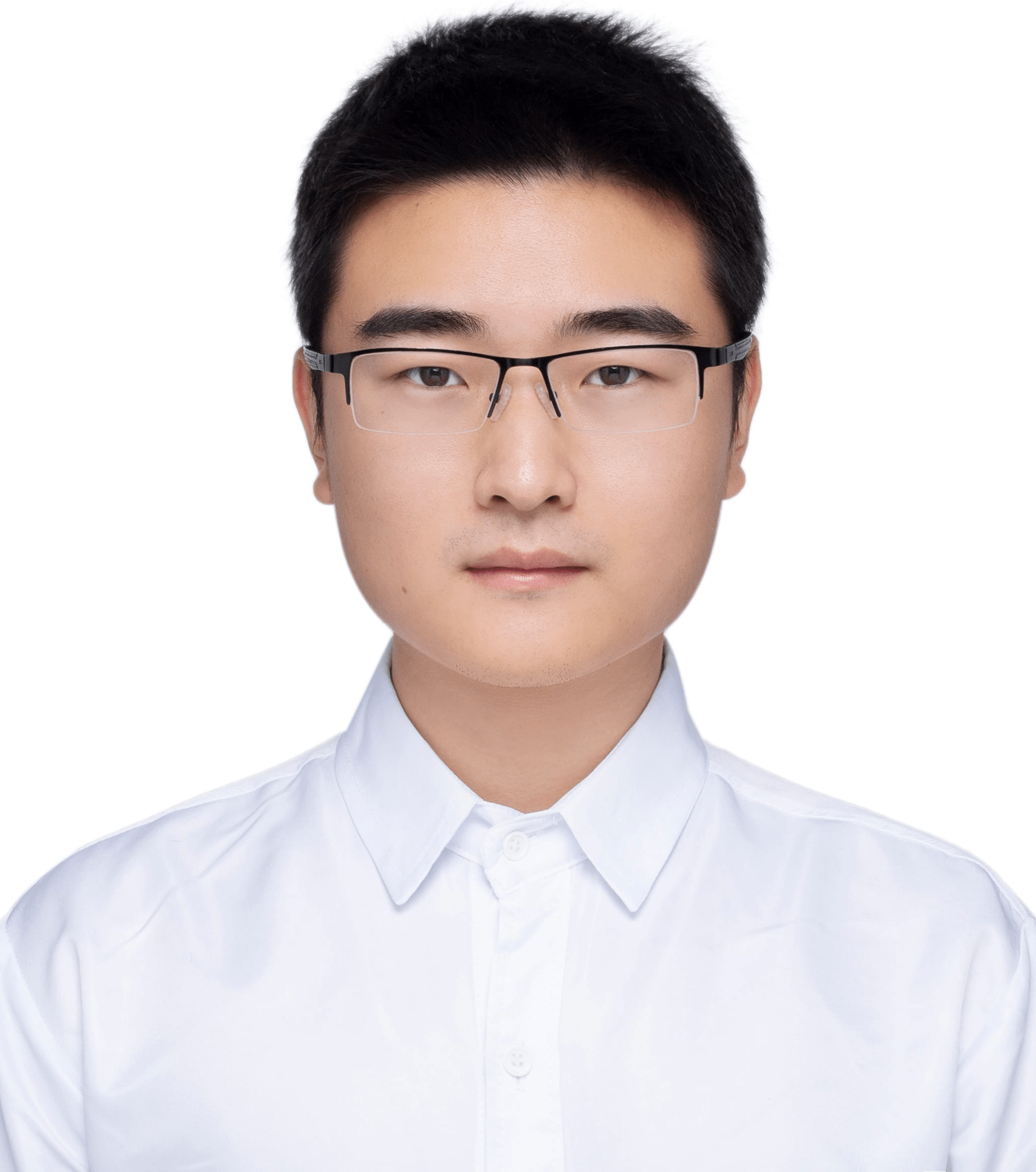}}]{Yuxuan Li} is currently a Ph.D. student at the Department of Computer Science, Nankai University, China.
  He graduated from University College London (UCL) with a first-class degree in Computer Science. He was the champion of the Second Jittor Artificial Intelligence Challenge in 2022, was awarded 2nd place in Facebook Hack-a-Project in 2019 and was awarded 2nd place in the Greater Bay Area International Algorithm Competition in 2022. His research interests include neural architecture design, and remote sensing object detection. 
\end{IEEEbiography}

\begin{IEEEbiography}[{\includegraphics[width=1.1in,height=1.35in,clip,keepaspectratio]{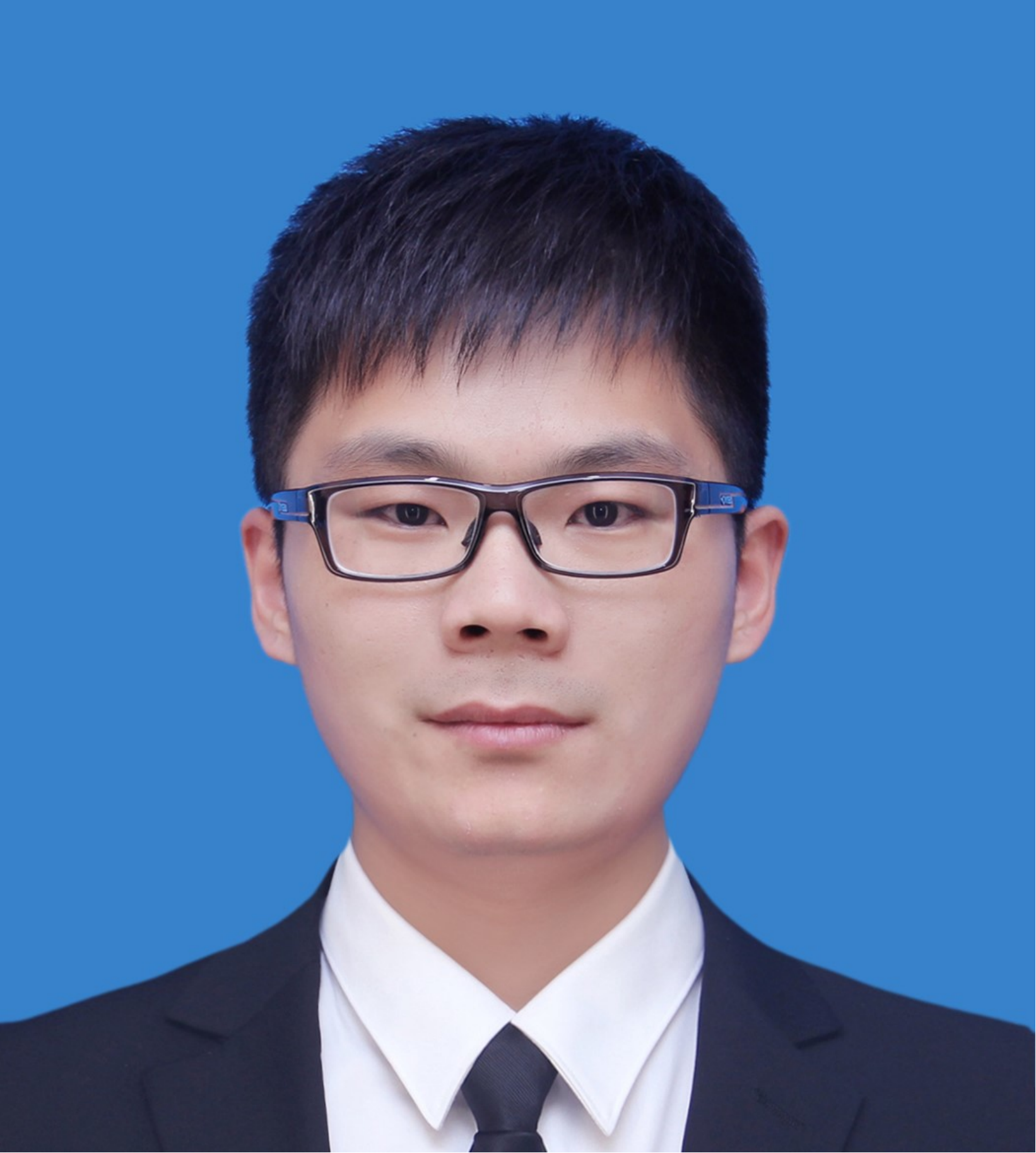}}]{Kang Ni (Member, IEEE) } received the M.S. degrees from Changchun University of Technology, Jilin, China, in 2016, and the Ph.D. degree from Nanjing University of Aeronautics and Astronautics, Jiangsu, China, in 2020.
  He is a Lecturer with the School of Computer Science, Nanjing University of Posts and Telecommunications, Nanjing, and also a member with the Jiangsu Key Laboratory of Big Data Security and Intelligent Processing, Nanjing. His research interests include machine learning, SAR image processing, and computer vision.
  \end{IEEEbiography}

\begin{IEEEbiography}[{
  \includegraphics[width=1.45in,height=1.3in,clip,keepaspectratio]{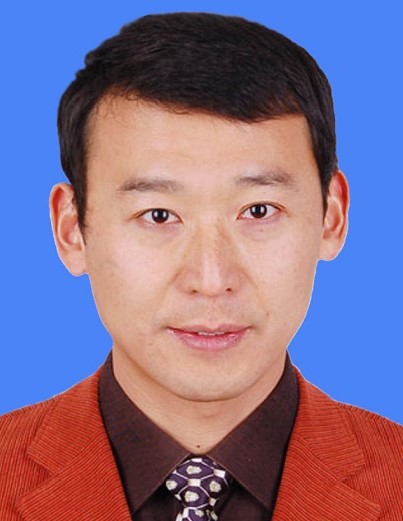}}]{Jian Yang} received the PhD degree from Nanjing University of Science and Technology (NJUST) in 2002, majoring in pattern recognition and intelligence systems. From 2003 to 2007, he was a Postdoctoral Fellow at the University of Zaragoza, Hong Kong Polytechnic University and New Jersey Institute of Technology, respectively. From 2007 to present, he is a professor in the School of Computer Science and Technology of NJUST. Currently, he is also a visiting distinguished professor in the College of Computer Science of Nankai University. He is the author of more than 300 scientific papers in pattern recognition and computer vision. His papers have been cited over 40000 times in the Scholar Google. His research interests include pattern recognition and computer vision. Currently, he is/was an associate editor of Pattern Recognition, Pattern Recognition Letters, IEEE Trans. Neural Networks and Learning Systems, and Neurocomputing. He is a Fellow of IAPR. 
\end{IEEEbiography}


\end{document}